%% file: main_arxiv_clvision.tex
\documentclass[10pt,twocolumn,letterpaper]{article}
\usepackage[pagenumbers]{cvpr}
\usepackage{graphicx}
\usepackage{amsmath}
\usepackage{amssymb}
\usepackage{stmaryrd}
\usepackage{booktabs}
\usepackage[dvipsnames]{xcolor}
\usepackage{flushend}

\usepackage[pagebackref=true,breaklinks=true,colorlinks,citecolor=NavyBlue,bookmarks=false]{hyperref}
\usepackage{multirow}
\usepackage{booktabs}
\makeatletter
\@namedef{ver@everyshi.sty}{}
\makeatother
\usepackage{tikz}
\usepackage{pgfplots}
\usepackage{wrapfig}
\pgfplotsset{compat=1.9}
\usepackage{subcaption}
\usepackage{makecell}
\usepackage{enumitem}
\usepackage{graphicx}
\usepackage{tabu}

\input{macros}

\usepackage[capitalize]{cleveref}
\crefname{section}{Sec.}{Secs.}
\Crefname{section}{Section}{Sections}
\Crefname{table}{Table}{Tables}
\crefname{table}{Tab.}{Tabs.}

\begin{document}

\title{CSG0: Continual Urban Scene Generation with Zero Forgetting}

\author{Himalaya Jain\textsuperscript{1}\footnotemark[1] \hspace{0.4cm} Tuan-Hung Vu\textsuperscript{1 }\footnotemark[1] \hspace{0.4cm} Patrick P\'erez\textsuperscript{1} \hspace{0.4cm} Matthieu Cord\textsuperscript{1,2}\\\\
\textsuperscript{1}Valeo.ai, Paris, France \hspace{0.8cm}
\textsuperscript{2}Sorbonne University, Paris, France
}
\maketitle

\begin{abstract}
    With the rapid advances in generative adversarial networks (GANs), the visual quality of synthesised scenes keeps improving, including for complex urban scenes with applications to automated driving. We address in this work a continual scene generation setup in which GANs are trained on a stream of distinct domains; ideally, the learned models should eventually be able to generate new scenes in all seen domains. This setup reflects the real-life scenario where data are continuously acquired in different places at different times. In such a continual setup, we aim for learning with zero forgetting, \IE, with no degradation in synthesis quality over earlier domains due to catastrophic forgetting. To this end, we introduce a novel framework that not only (i) enables seamless knowledge transfer in continual training but also (ii) guarantees zero forgetting with a small overhead cost. While being more memory efficient, thanks to continual learning, our model obtains better synthesis quality as compared against the brute-force solution that trains one full model for each domain. Especially, under extreme low-data regimes, our approach outperforms the brute-force one by a large margin.
\end{abstract}

\section{Introduction}
\input{CLVISION_intro}

\section{Related work}
\input{CLVISION_relatedwork}

\section{CSG0  for continual scene generation}
\input{CLVISION_method}

\section{Experiments}
\input{CLVISION_experiments}

\section{Conclusion}
\input{CLVISION_conclusion}

\appendix
\section{Appendix}
Figures~\ref{fig:qual_res_idd_suppmat} and~\ref{fig:qual_res_mapillary_suppmat} show more qualitative results of our CSG0 models.
In Table~\ref{tbl:csg0_mappings}, we detail the classes used in continual urban scene generation; In each dataset, we show the class names, the original class Ids and the corresponding Ids in the continual setups.
{\small
\bibliographystyle{ieee_fullname}
\bibliography{egbib}
}
\input{CLVISION_appendix}

\end{document}

%% file: macros.tex
\usepackage{amsmath}
\usepackage{amssymb}
\usepackage{dsfont}
\usepackage{bm}
\usepackage{xcolor,soul,colortbl}
\usepackage{array}
\usepackage{multirow}
\usepackage{subcaption}
\usepackage{dashrule}
\usepackage{tabularx}
\usepackage{upgreek}
\usepackage{graphicx}
\usepackage{enumitem}
\usepackage{booktabs}
\usepackage{stmaryrd}

\usepackage{tikz}
\usepackage{pgfplots}
\usepackage{wrapfig}
\pgfplotsset{compat=1.9}
\usepackage{makecell}

\newcommand{\IE}{\textit{i}.\textit{e}.\@}
\newcommand{\EG}{\textit{e}.\textit{g}.\@}

\newcommand{\paragr}[1]{\noindent\textbf{#1}~} 
\newcommand{\parag}[1]{\smallskip\noindent\textbf{#1}~}

\newcommand{\revcorr}[2]{\textcolor{black}{#2}}      
\newcommand{\NEW}[1]{\textcolor{black}{#1}}

\definecolor{CA}{HTML}{E6E6FA}
\definecolor{CS}{HTML}{FFEFD5}
\definecolor{CE}{HTML}{FFC0CB}

\definecolor{tcolor}{RGB}{219,50,54}
\definecolor{scolor}{RGB}{72,133,237}

\definecolor{gGreen}{RGB}{15,157,88}
\definecolor{gBlue}{RGB}{66,133,244}
\definecolor{gOrange}{RGB}{230, 145, 56}

\definecolor{cidd}{RGB}{220,20,60}
\definecolor{cnot_idd}{RGB}{174,174,174}
\definecolor{cmapillary}{RGB}{107,142,35}
\definecolor{cnot_mapillary}{RGB}{174,174,174}

\newcommand{\mv}[1]{\ensuremath{\bm{#1}}} 
\newcommand{\mm}[1]{\ensuremath{\bm{#1}}} 
\newcommand{\rtb}[1]{\rotatebox{90}{#1}}   

\newcolumntype{C}[1]{>{\centering\arraybackslash}p{#1}}	
\newcolumntype{L}[1]{>{\raggedright\arraybackslash}p{#1}}
\newcolumntype{R}[1]{>{\raggedleft\arraybackslash}p{#1}}

\newcommand{\GtoI}{GTA5$\shortrightarrow$IDD}
\newcommand{\GtoM}{GTA5$\shortrightarrow$Mapillary}
\newcommand{\CtoI}{Cityscapes$\shortrightarrow$IDD}
\newcommand{\CtoM}{Cityscapes$\shortrightarrow$Mapillary}
\newcommand{\GtoItoM}{GTA5$\shortrightarrow$IDD$\shortrightarrow$Mapillary}
\newcommand{\CtoItoM}{Cityscapes$\shortrightarrow$IDD$\shortrightarrow$Mapillary}
\newcommand{\E}{\mathbb{E}}

%% file: CLVISION_intro.tex
\footnotetext{$\ast$ Equal contribution} 
Visual scene synthesis with generative adversarial networks (GANs) conditioned on input semantic segmentation masks is progressing fast. 
Since the early work of Pix2Pix~\cite{pix2pix_isola2017image}, many architecture designs and learning strategies~\cite{pix2pixhd_wang2018high,spade_park2019semantic,liu2019learning,oasis_schonfeld2020you} were proposed to push forward the synthesis quality, making generated images look more and more realistic.
However, most existing works are limited to a single-domain setting, \IE, once trained, the GAN can only generate images close to the distribution of the training domain.
Recent works propose techniques for fine-tuning a pre-trained GAN~\cite{wang2018transferring,mo2020freeze}, training specific parameters \cite{shahbazi2021efficient} or identifying the most favourable regions on the learned manifold~\cite{wang2020minegan}, to help transfer learning to new domains.

\begin{figure}
	\small
	\centering
	\includegraphics[width=0.48\textwidth]{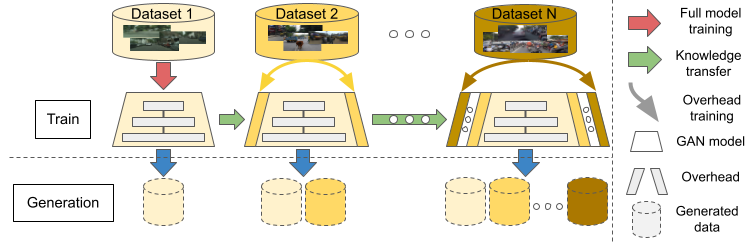}
	\caption{\small\textbf{Overview of the proposed framework}. Our continual setup for urban-scene generation involves a stream of datasets, with GANs trained from one dataset to another. Our framework makes use of the knowledge learned from previous domains and adapts to new ones with a small overhead. Best viewed in color.}
	\label{fig:teaser}
	\vspace{-0.6cm}
\end{figure}

We here tackle the task of continual urban-scene generation for multiple domains using GANs (see overview in Fig.\,\ref{fig:teaser}).
We address a realistic scenario that takes into account the continual property of driving data acquisition: data are continually collected from different places.
Such a reality asks for efficient mechanisms to continuously extend generative models, which were previously trained, to newly collected datasets.
The straight-forward solution is to fine-tune the current model using the new data.
However this solution greatly suffers from the ``catastrophic forgetting'' phenomena, often met with data-driven models~\cite{seff2017continual,wu2018memory,zhai2019lifelong}.

In this work, we aim for preserving at best the synthesis quality in all domains, \IE, we want to avoid catastrophic forgetting altogether, seeking instead no- or zero-forgetting.
To this end, one could train a separate model for each domain, at the expense of significant memory consumption when the number of domains grows. 
We argue that, although urban-scene datasets look different in color/texture and even have different label spaces, they share lots of structure, \EG, scene arrangements and shapes of objects in shared classes.
Therefore, given a GAN pre-trained on a related domain, it should be unnecessary to have all model's weights learned again for the new domain; instead, only a minimal set of parameters should need learning. 
Based on this rationale, we propose a novel framework for continual scene generation with zero-forgetting. 
Building around the idea of modulating network weights, 
we approach the continual task with care, analyze the requirements for suitable architecture designs and learning strategies to handle the extension of the label space in the new domains, with minimal overhead and large adaptability.
Effectively, we seek a good trade-off between the complexity overhead required by  zero-forgetting and the image synthesis quality.
In brief, the main contributions of this paper are as follows:
\begin{itemize}[leftmargin=10pt,topsep=2pt,itemsep=2pt,parsep=0pt,partopsep=0pt]
	\item We address the novel task of continual learning of GAN for semantic scene generation, where each new domain comes with new semantic classes and new visual styles. To the extent of our knowledge, this is the first work addressing continual scene generation.
	\item We propose a modular approach, named CSG0, to tackle the problem and evaluate the contribution of each module in the context of urban scenes.
	\item We show in various continual setups, covering both synthetic-to-real and real-to-real scenarios, that CSG0 outperforms state-of-the-art models trained on individual domains. 
	\item We demonstrate the merit of CSG0 in low-data regimes in which learning is done with only a few tens of samples.
\end{itemize}

%% file: CLVISION_relatedwork.tex
\parag{Scene generation with GANs.} Image synthesis with GANs conditioned on semantic maps has progressed significantly.
Pix2Pix~\cite{pix2pix_isola2017image} is the first work to address this problem, using an encoder-decoder generator with PatchGAN discriminator. Pix2PixHD~\cite{pix2pixhd_wang2018high} proposes a coarse-to-fine generator architecture and multiple PatchGAN discriminators to generate high-resolution images.
SPADE \cite{spade_park2019semantic} proposes a spatially-adaptive normalization layer that modulates the feature maps of the generator. 
The modulating parameters are predicted based on the input semantic map. 
This explicit use of the input semantic map to control the structure of the generated images improves the fidelity to it. 
Similarly, to have an explicit impact of the semantic map, CC-FPSE~\cite{liu2019learning} proposes to predict convolutional kernels of the generator conditioned on it.
OASIS~\cite{oasis_schonfeld2020you} proposes a segmentation-based discriminator and shows that only the adversarial loss is sufficient to get high-fidelity generation unlike previous works that require a perceptual loss. OASIS generator extends the SPADE layers to take jointly noise and semantic map as input; this enhances the impact of noise and thus the diversity of the generated images. In our work, we use OASIS as our base framework for urban-scene generator and explore various directions for continual learning with it.

\paragr{Continual learning for GANs.}
Continual learning for GANs was first addressed in~\cite{seff2017continual}, where elastic weight consolidation (EWC)~\cite{kirkpatrick2017overcoming} is used to avoid catastrophic forgetting.
LifelongGAN~\cite{zhai2019lifelong} proposes to use knowledge distillation from the previous generator to the current one to address forgetting of the previous tasks. 
Memory replay GAN~\cite{wu2018memory} uses the generated images of the previous tasks with the current task images to train the current generator. In~\cite{zhai2019lifelong, wu2018memory}, all weights of the generator are fine-tuned, thus still suffering from some forgetting. 
In Hyper-lifelongGAN~\cite{zhai2021hyper}, convolutional filters of the generator are decomposed into the dynamic task-specific base filters and a deterministic generic weight matrix. Knowledge distillation is used to ensure good performance on the previous tasks.
Recent works proposed to learn additional task-specific parameters while keeping the remaining generator frozen to preserve the performance on the previous tasks.
Piggybank GAN~\cite{zhai2020piggyback} learns new task-specific filters. For the current task, these filters are used in combination with filters from a bank of filters learned on the previous tasks.
GAN memory~\cite{cong2020gan} proposes to learn task-specific weight modulation parameters, that is, task-specific mean and standard deviation of the weight matrices of the generator.

\NEW{The existing continual GAN approaches are addressing continual learning for either unsupervised~\cite{cong2020gan}, class-conditioned~\cite{wu2018memory, zhai2019lifelong, cong2020gan} or image-conditioned~\cite{zhai2019lifelong, zhai2020piggyback, zhai2021hyper} GANs.
In this case, for each new task, the conditioning input domain (set of classes or images) is completely replaced by a new domain; thus there is no overlap or sharing between the input domains across the tasks.
In this work,}
we focus on continual learning for the task of semantic scene generation.
This brings some new aspects to continual GAN learning.
In particular, the label space of the previous tasks should be extended rather than replaced when accommodating a new incoming domain.
To our best knowledge, our work is the first to address continual learning for semantic scene generation. As will be explained in the next section, we take the no-forgetting approach where we only learn some task-specific new parameters while preserving the performance on the previous tasks.

\paragr{Fine-tuning GANs.} Fine-tuning refers to continuing on a new target dataset the training of a pre-trained model. The objective is to achieve best performance on the new domain or task, regardless of catastrophic forgetting.
The first work to propose GAN fine-tuning is~\cite{wang2018transferring}, which shows that a pre-trained generator can indeed be fine-tuned on a new dataset, thus requiring less data and learning iterations.
FreezeD~\cite{mo2020freeze} proposes to freeze a few initial layers of the pre-trained discriminator while fine-tuning the rest on a new dataset. 
\cite{noguchi2019image} learns new BatchNorm parameters to fine-tune the generator on a very small dataset. 
\cite{shahbazi2021efficient} proposes to learn class-specific BatchNorm parameters by using knowledge from BatchNorm parameters of the pre-trained conditional GAN.
MineGAN~\cite{wang2020minegan} learns a miner network that produces latent codes for the pre-trained generator so as to gear it toward the new target distribution. In the second stage, all the networks --the miner, the generator and the discriminator-- are fine-tuned on the target dataset. 

While GAN fine-tuning is important, it is not directly applicable to continual learning without forgetting, which is our goal here for semantic urban-scene generation.

%% file: CLVISION_method.tex
We address the task of continual scene generation, conditioned on input semantic segmentation masks.
On a continuous stream of $N$ datasets, GAN training is done from one dataset to another.
At inference, the main goal is to purposely synthesize images coming from any of the $N$ domains.
Starting from a GAN model pre-trained on previous domains, we want to reuse most of the learned weights and extend the model with small overhead (i) to leverage the knowledge learned from the previous domains and (ii) to handle the new domain with new classes using the added parameters.

To this end, several challenges must be overcome.
First, as we are dealing with new classes in the new domain, the continual model must be re-designed so that it can accept those classes as inputs.
Second, we need an efficient mechanism to reuse most of the parameters learned from the previous domains while allowing sufficient degrees of freedom so that the continual model can adapt to the new domain with a different style.
Furthermore, by default, most GAN networks adopt Batch Normalization to stabilize  training, which may not be ideal for our task where we want a more explicit control and adaptation of the domain's ``style''.
\revcorr{Last but not least, knowing that the old and new classes may coexist in the same scene, we cannot separate learning of the old and the new; instead, it must be done jointly. 
This raises a risk of negative interference between the old and new parameters.
We thus need a good strategy to mitigate such a risk.}{}
In what follows, we present all technical details of the base generative model, in Section~\ref{sec:oasis}, and our proposed continual strategies, in Section~\ref{sec:csg0}.

\subsection{GAN OASIS framework}
\label{sec:oasis}
We use the recent scene generator OASIS~\cite{oasis_schonfeld2020you} as our base framework.
As input it takes a 3D noise concatenated with one-hot segmentation maps. 
The input is fed at various levels using SPADE blocks~\cite{spade_park2019semantic}.
The generator $G$ has six ResNet blocks, each block is structured as SPADE-Conv-SPADE-Conv with a skip connection. The key idea of OASIS training lies in the design of the discriminator $D$, which is a segmentation-like network.
This discriminator is trained not only to discriminate real/fake pixels, but also to classify real pixels into the correct semantic classes. 

Denoting $C$ the number of semantic classes at hand, $D(\cdot)$ thus outputs maps of size  $H{\times}W{\times}(C{+}1)$ given an input image $\mv x$ of size $H{\times}W$; Accordingly, the generator $G(\cdot)$ produces images of the same size, given a binary semantic tensor $\mm S$ of size $H{\times}W{\times}C$ and a noise vector $\mv z$ (turned into a 3D tensor by replication over the pixel grid) as inputs.     

For training, we use the objectives proposed in OASIS, that is, an adversarial loss for the generator and a segmentation loss (with an additional \textit{fake} class) combined with LabelMix regularization for the discriminator. We provide below an abridged description of the loss terms; for more details readers could refer to~\cite{oasis_schonfeld2020you}.

Let $\mm{S}$ denote a one-hot input segmentation map, where $\mm{S}_{i,j,c}\in\{0,1\}$ is non zero only if pixel $(i, j)$ is labelled as class $c$. The generator is trained to minimize,
\begin{equation}
    \mathcal{L}_G = - \E_{(\mv{z}, \mm{S})} \sum_{c=1}^{C} \alpha_c \sum_{i=1}^{H}\sum_{j=1}^{W} \mm{S}_{i,j,c} \log D(G(\mv{z}, \mm{S}))_{i,j,c},
\end{equation}
where the $\alpha_c$'s are class-balancing weights, defined as inverse class frequencies, and expectation is approximated over pairs of random noise vectors and true semantic maps.

The discriminator loss is defined as 
\begin{multline}
    \mathcal{L}_D = - \E_{(\mv{x}, \mm{S})} \sum_{c=1}^{C} \alpha_c \sum_{i=1}^{H}\sum_{j=1}^{W} \mm{S}_{i,j,c} \log D(\mv{x})_{i,j,c} \\
    - \E_{(\mv{z}, \mm{S})} \sum_{i=1}^{H}\sum_{j=1}^{W} \log D(G(\mv{z}, \mm{S}))_{i,j,c=N+1},
\end{multline}
that is, the combination of the $C$-class cross-entropy loss for real pixels and the binary cross-entropy loss for fake ones; the first expectation is taken over real images and associated ground-truth segmentation maps. 

The discriminator is further trained with LabelMix regularization. A LabelMix image
is formed by mixing real and generated images associated with the same ground-truth segmentation map according to a binary mask $\mm{M}$, \IE, $\operatorname{LabelMix}(\mv{x}, G(\mv{z}, \mm{S}), \mm M) = \mm M \mv{x} + (1-\mm M) G(\mv{z}, \mm{S})$, where operations are meant entry-wise. The discriminator is regularized to be pixel-wise consistent in its prediction on LabelMix images and on corresponding real and generated images before mixing. 

While we use OASIS as the base for our continual learning scene generation framework and its evaluation, note that the approach is not limited to OASIS. 

\begin{figure*}
    \centering
    \includegraphics[width=\textwidth]{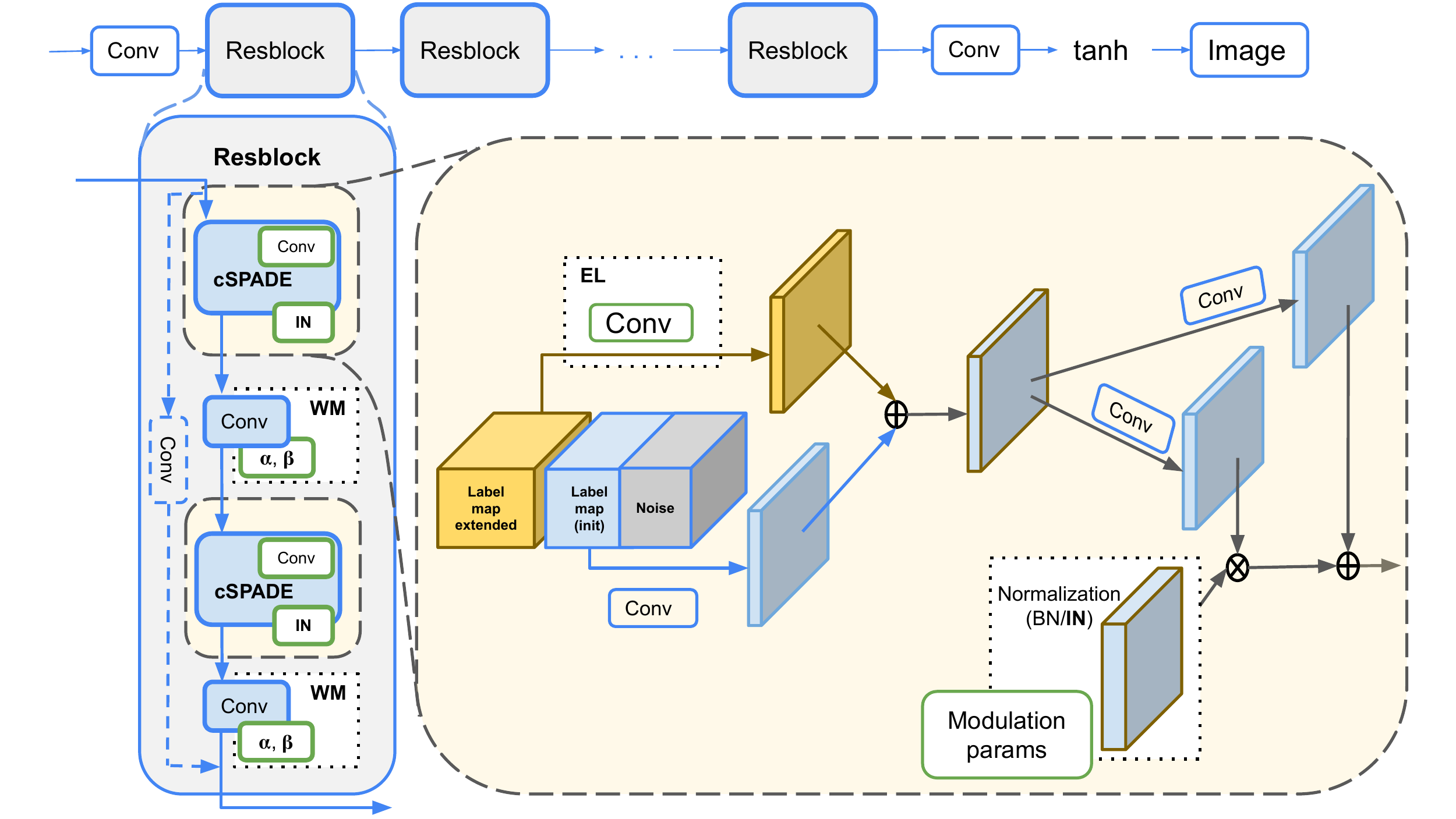}
    \caption{\small \textbf{Proposed framework for continual semantic image generation}.
    Starting off from an OASIS model shown on top, we propose new designs for continual learning.
    All modules drawn in blue are frozen during continual training, only green ones are learned.
    Yellow blocks stand for the newly introduced input in the new domain as well as its corresponding features.
    \revcorr{The ``Mask'' blocks in magenta illustrate the masking mechanism to prevent negative interference between ``old'' and ``new'' blocks.}{}
    Details are given in Section~\ref{sec:csg0}.
    }
    \label{fig:approach}
\end{figure*}

\subsection{CSG0 Model}
\label{sec:csg0}
We now detail our continual strategies and introduce the continual scene generation model with zero-forgetting, namely CSG0.
Figure~\ref{fig:approach} illustrates our architecture designs.
We start CSG0 from the OASIS model $G_o$ pre-trained on previous datasets with $C_o$ semantic classes. 

\parag{Extending the input label space.}
We now consider a new domain with $C_n$ \textit{new} classes. 
To accommodate it, we need to re-design the input layers of the generator to make them ingest semantic masks with $C_o{+}C_{n}$ classes. 
To this end, we modify the SPADE blocks in OASIS to accept the new 3D semantic tensors composed of $(C_o{+}C_n)$-dim one-hot vectors. 
Originally, the first convolutional layer (\texttt{conv}) of SPADE contains $1024$ kernels of size $C_o{\times}\,3\,{\times}\,3$.
For new classes, we introduce a new \texttt{conv} layer, named ``EL'', standing for ``Extented Labels'', with similarly $1024$ kernels, now of size $C_{n}{\times}\,3\,{\times}\,3$.

We then split the input 3D tensor into the ``new'' and ``old'' 3D tensors, corresponding to the new and old classes.
Effectively, the two split tensors are composed of one-hot vectors of $C_o$-dim and $C_n$-dim respectively. 
Areas of old classes are encoded by zero vectors in the new 3D tensor, and vice versa.
The two split 3D tensors are fed into corresponding \texttt{conv} layers, whose outputs are then summed up to obtain one final output.
This new SPADE block, modified for continual learning, will be referred to as ``cSPADE'', to distinguish it from the original one. 
The structure of this block is represented in Figure~\ref{fig:approach}, where the new input 3D tensor and the output of the new EL \texttt{conv} are highlighted in yellow, in contrast to the original 3D tensor and \texttt{conv} layer which are in blue.

\parag{Weight modulation.}
Weights of the convolutional layers encode most of the knowledge from the previous domains.
We want to retain important information encoded in these weights while allowing certain adaptability to handle the new domain.
To this end, we adopt the mAdaFM technique introduced in~\cite{cong2020gan}, which helps modulate the ``style'' of the \texttt{conv} layer.
The technique brings to \texttt{conv} layers the idea of statistics modulation used for style transfer~\cite{dumoulin2016learned,huang2017arbitrary}; intuitively, \texttt{conv} weights are regarded as network's ``features'' with domain-invariant and domain-specific parts.
The key idea is to keep the domain-invariant part frozen while allowing learning in the domain-specific part, which eventually results in weight adaptation to the new domain.

In detail, given a \texttt{conv} layer with $C_{\text{in}}$ input and $C_{\text{out}}$ output channels, weight matrix $\mathbf{W}\,{\in}\,\mathbb{R}^{C_{\text{out}}{\times}C_{\text{in}}{\times}K \times K}$, bias vector $\bm{b}\,{\in}\,\mathbb{R}^{C_{\text{out}}}$ and kernel size $K$, we modulate those parameters in the new \texttt{conv} by defining the new weight matrix $\widehat{\mathbf{W}}$ and bias vector $\hat{\bm{b}}$ as: 
\begin{equation}
    \widehat{\mathbf{W}} = \bm{\alpha}\odot\frac{\mathbf{W}-\mathbf{M}}{\mathbf{S}} + \bm{\beta},\,\,\,\,\,\,\,\,\,\hat{\bm{b}} = \bm{b} + \bm{b}_{\text{conv}},
\end{equation}
where: $\mathbf{M}$ and $\mathbf{S} \,{\in}\,\mathbb{R}^{C_{\text{out}}{\times} C_{\text{in}}}$ with entries $\mathbf{M}_{i,j}$ and $\mathbf{S}_{i,j}$ being the mean and standard deviation respectively of $(\mathbf{W}_{i,j,p,k})_{1\leq p,k \leq K}$; $\odot$ is the Hadamard product; $\bm{\alpha}\,{\in}\,\mathbb{R}^{C_{\text{out}}\times C_{\text{in}}}$ (scale), $\bm{\beta}\,{\in}\,\mathbb{R}^{C_{\text{out}}\times C_{\text{in}}}$ and $\bm{b}_{\text{conv}}\,{\in}\,\mathbb{R}^{C_{\text{out}}}$ (shift) are new domain-specific parameters.
The learning process on the new domain only updates these domain-specific parameters, while leaving the base parameters $\mathbf{W}$ and $\bm{b}$ untouched. 
In the Resblock of CSG0, we apply weight modulation for the two \texttt{conv} layers coming after each cSPADE block.
The learnable parameters $\bm{\alpha}$ and $\bm{\beta}$ are highlighted in the green boxes in Figure \ref{fig:approach}.

\parag{Instance normalization.}
Batch normalization (BN) and modulation of the activation in the generator have shown to have a strong influence on the style of the generated images.  This is also especially true for Instance Normalization (IN)~\cite{ulyanov2016instance}.
Inspired by this observation that IN provides better control of the style, we extend our minimal EL setup by replacing BN with IN with affine transform. Note that using IN with affine transformation does not require any additional parameters compared to BN. The normalization block in Figure \ref{fig:approach} shows this module in the full framework. With `Modulation params' in the figure, we mark that these are dataset-specific parameters which are used to modulate the activation in the normalization layer.

\revcorr{\parag{Masking to prevent negative transfer.}
As discussed earlier, one risk of having learned parameters of old classes and new parameters of new classes both in place is to get negative interference. 
Indeed, as the EL \texttt{conv} regards old-class areas as `unknown' (zero vectors), outputs in such locations are not reliable for contributing to pixel generation of old classes. 
A similar thing happens for the new-class areas within the original first \texttt{conv}.
We therefore propose a masking mechanism to explicitly prevent such an issue from happening. 
From the input segmentation maps, we first mark the areas of new classes in a binary mask. We then block the influence from one area to another: element-wise multiplication between this mask and the output of EL \texttt{conv} layers zeroes out the areas corresponding to old classes; conversely, we zero out new-class areas in the output of the original first \texttt{conv} layers by element-wise product with the complementary mask. This masking mechanism is illustrated in the `Mask' dashed box in Figure~\ref{fig:approach}.
}{}

%% file: CLVISION_experiments.tex
\subsection{Experimental details}

\begin{table*}[ht!]
    \setlength\heavyrulewidth{0.25ex}
    \setlength{\tabcolsep}{6pt}
    \aboverulesep=0ex
    \belowrulesep=0ex
	\centering
	\small
    \begin{minipage}[c]{0.54\textwidth}
        \centering
        \begin{tabular}{@{}L{0.1cm}@{}@{}L{2.9cm}@{}@{}C{1.2cm}@{}@{}C{1.2cm}@{}@{}C{0.86cm}@{}C{0.86cm}@{}@{}C{0.86cm}@{}C{0.86cm}@{}}
            \toprule
            \multicolumn{2}{c}{\multirow{2}{*}{\small Model}} &  \multirow{2}{*}{\parbox{1.2cm}{\small \centering New\\ params}} &  \multirow{2}{*}{\parbox{1.2cm}{\small \centering Total\\ params}}	& 	\multicolumn{2}{c}{\small G $\rightarrow$ I} &	\multicolumn{2}{c}{\small C $\rightarrow$ I} \\	 
            \cmidrule(lr){5-6}\cmidrule(lr){7-8}
            \multicolumn{1}{c}{}& & & &	FID & mIoU	&	FID & mIoU		\\	
            \midrule
            \multicolumn{1}{c}{\multirow{3}{*}{\rtb{CSG0}}}&\,\,cSPADE &	{0.3M} & 71.5M	&	63.1 & 28.6 &	75.2 & 21.9 \\
            &\,\,cSPADE$\,{+}\,$IN & {0.3M} & 71.5M &	\textcolor{gGreen}{50.7} & \textcolor{gGreen}{32.2} &	\textcolor{gBlue}{55.1} & \textcolor{gBlue}{29.3} \\
            &\,\,cSPADE$\,{+}\,$IN$\,{+}\,$WM &	10.8M &  82.0M &	\textbf{\textcolor{gGreen}{39.1}} & \textbf{\textcolor{gGreen}{39.4}}&	\textbf{\textcolor{gBlue}{38.4}} & \textbf{\textcolor{gBlue}{35.2}}	\\
            \midrule
            &\,\,OASIS (I)~\cite{oasis_schonfeld2020you} &	71.4M & 142.6M	&	\textcolor{gOrange}{55.3} & \textcolor{gOrange}{41.0}  &	\textit{idem} & \textit{idem} \\
            \bottomrule
        \end{tabular}
    \end{minipage}
    \hfill
    \begin{minipage}[c]{0.43\textwidth}
        \centering
        \includegraphics[width=\textwidth]{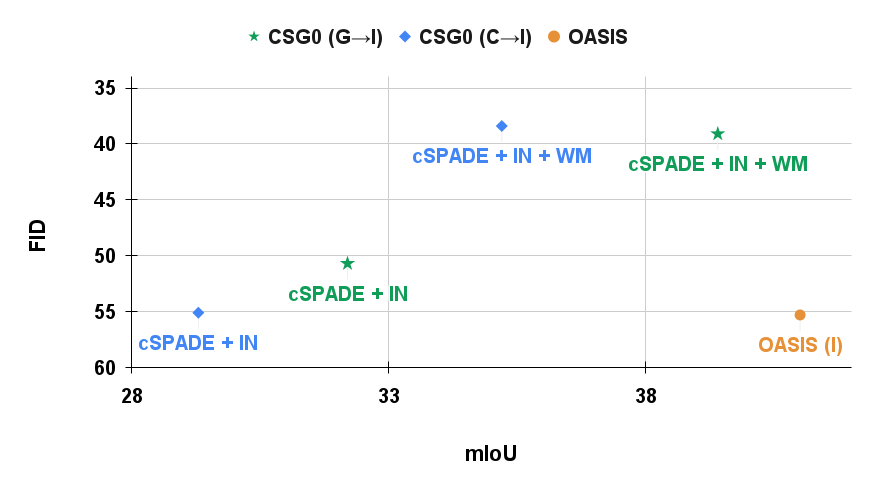}
    \end{minipage}
    \vspace{-0.3cm}
    \caption{\small\textbf{Performance on IDD of continual models}. All continual models are initialized from an OASIS model, trained either on GTA5 (G) or Cityscapes (C), having 72M parameters. Table to the \textit{left}: the first four rows report results of ablated CSG0 models on different modules: cSPADE stands for the use of EL \texttt{conv} layer, IN and WM stand for InstanceNorm and `Weight Modulation' strategies; the last row is for the `OASIS (I)' model only trained on IDD.
    Some colored results in the table are plotted in the figure to the \textit{right} with the corresponding colors.
    The OASIS model, though having good mIoU score, obtains worse FID as compared to CSG0 models.
    Such results are matched with synthesis quality of examples shown in Figure~\ref{fig:qual_res_idd}.}
    \label{tbl:GC2I}
\end{table*}

\begin{figure*}
    \small
    \hspace{-0.4cm} 
    \begin{tabular}{cccc}
    cSPADE&cSPADE$\,{+}\,$IN& cSPADE$\,{+}\,$IN$\,{+}\,$WM & OASIS(I) \\
    \includegraphics[width=0.24\textwidth]{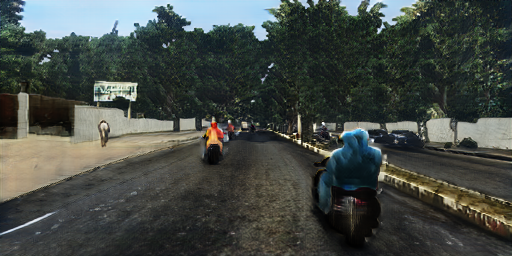} \hspace{-5pt} &
    \includegraphics[width=0.24\textwidth]{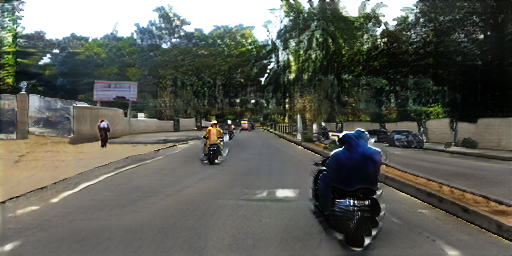} \hspace{-5pt} &
    \includegraphics[width=0.24\textwidth]{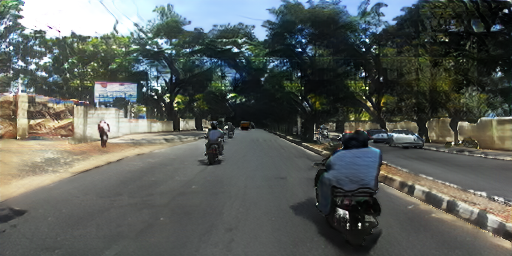} \hspace{-5pt} &
    \includegraphics[width=0.24\textwidth]{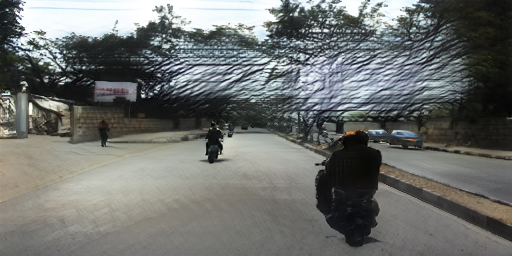}\\
    
    \includegraphics[width=0.24\textwidth]{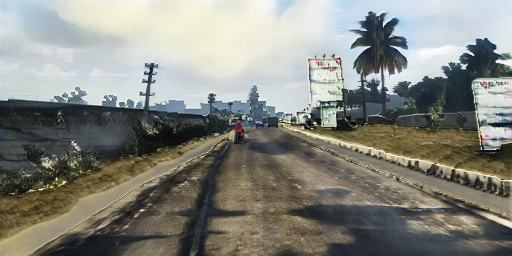} \hspace{-5pt} &
    \includegraphics[width=0.24\textwidth]{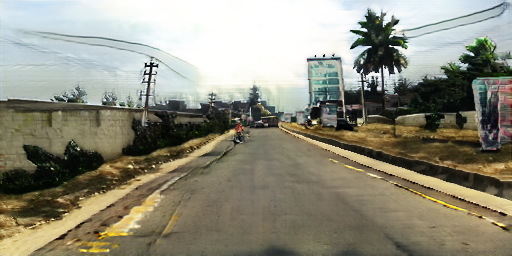}  \hspace{-5pt} &
    \includegraphics[width=0.24\textwidth]{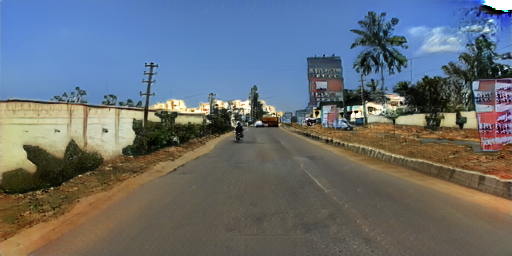} \hspace{-5pt} & 
    \includegraphics[width=0.24\textwidth]{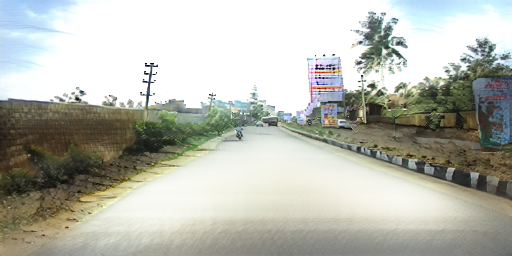}\\
    
    \includegraphics[width=0.24\textwidth]{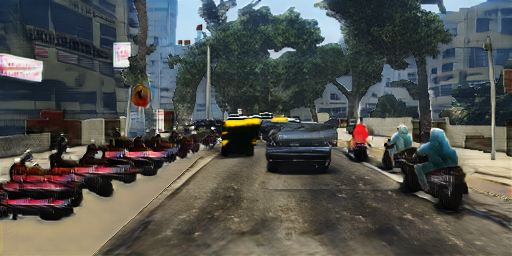} \hspace{-5pt} &
    \includegraphics[width=0.24\textwidth]{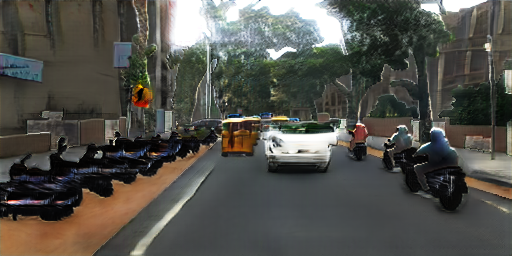} \hspace{-5pt} &
    \includegraphics[width=0.24\textwidth]{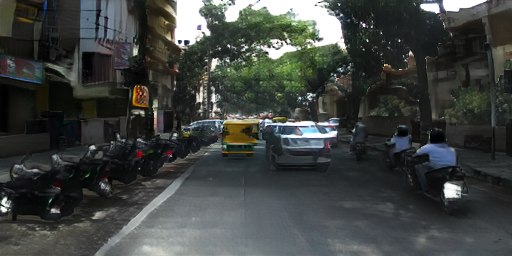} \hspace{-5pt} &
    \includegraphics[width=0.24\textwidth]{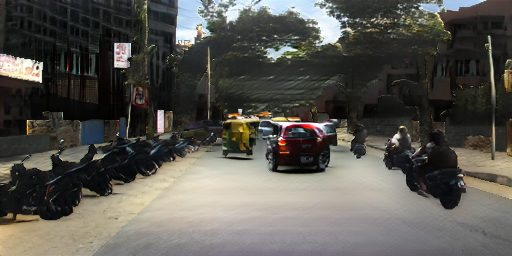}
    \end{tabular}
    \caption{\small\textbf{Qualitative results on IDD}. Each row visualizes images synthesized from the same input semantic segmentation mask using different models. All continual models are initialized from the OASIS model that was trained on GTA5. Results of the basic CSG0 model that only has the vanilla cSPADE still retain the color and texture of GTA5. The instance normalization strategy helps facilitate style learning, resulting in images with more IDD-like tone. Having ``Weight Modulation'' (WM) improves further the quality. The `OASIS (I)' model introduces visible artifacts, especially in ``tree'' and ``sky'' areas; also textures are quite repetitive. Our full CSG0 model produces the most realistic images with natural color tone and texture.
    Best viewed in color.}
    \label{fig:qual_res_idd}
\end{figure*}

\parag{Continual set-ups.}
In this work, we focus on generating urban scenes, which leaves us a few dataset choices.
Our selection criterion is to have datasets that have rather different visual styles and ``private'' semantic classes that do not exist in others.
The final shortlist contains GTA5~\cite{richter2016playing}, composed of synthetic urban scenes, Cityscapes~\cite{cordts2016cityscapes}, collected in Germany and Switzerland, Indian Driving Dataset (IDD)~\cite{varma2019idd}, acquired in India, and Mapillary~\cite{neuhold2017mapillary} with scenes from all around the world. 
Addressed setups cover both synthetic-to-real and real-to-real scenarios, each with different numbers of domains.
We address in our experiments the six following continual training sequences of 2 or 3 datasets:
\begin{itemize}[leftmargin=10pt,topsep=2pt,itemsep=2pt,parsep=0pt,partopsep=0pt]
\item synth-to-real:\,\GtoI{} and \GtoM 
\item real-to-real:\,\CtoI{} and \CtoM
\item synth-to-real-to-real:\,\GtoItoM
\item real-to-real-to-real:\,\CtoItoM.
\end{itemize}

\parag{Continual training.}
In each continual set-up, we start off with a vanilla OASIS model trained on the first dataset.
Pre-trained weights of this OASIS are used to initialize the continual CSG0 models.
All new CSG0 parameters, which are dedicated to the next datasets, are newly initialized.

\parag{Evaluation protocol.}
In this work, we are mainly concerned with transferring with minimal overhead the knowledge learned from previous datasets to the current one, so as to achieve good synthesis quality.  
The most important is thus finding a sweet spot in the trade-off between complexity overhead and image quality. 
The former is measured by the number of newly introduced parameters, and the total number of parameters needed for generation in all domains. 
To assess the latter, we regard FID~\cite{heusel2017gans} as the main metric, similarly to~\cite{pix2pixhd_wang2018high,spade_park2019semantic,liu2019learning,oasis_schonfeld2020you}.
Following~\cite{oasis_schonfeld2020you}, we additionally report the mean intersection-over-union (mIoU) obtained when testing a pre-trained PSPNet~\cite{zhao2017pyramid} segmentation model on the generated data; this metric is named `GAN-test'.
We note the difficulty of quantitatively assessing the synthesis quality: on some metrics like GAN-test, good scores do not always match visual quality.
In this regard, FID appears to be the most meaningful metric.

\subsection{Main results}
\begin{figure*}
    \hspace{-0.2cm}
    \small
    \begin{tabular}{cccc}
    cSPADE&cSPADE$\,{+}\,$IN&cSPADE$\,{+}\,$IN$\,{+}\,$WM& OASIS(Mapillary)\\

    \includegraphics[ width=0.24\textwidth]{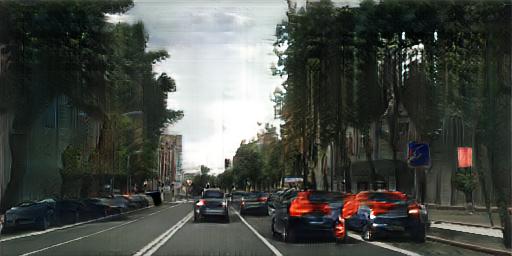}\hspace{-7pt} &
    \includegraphics[ width=0.24\textwidth]{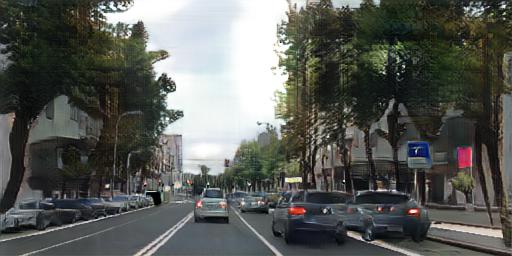}\hspace{-7pt} &
    \includegraphics[ width=0.24\textwidth]{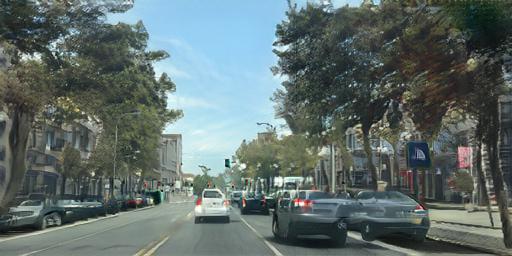}\hspace{-7pt} &
    \includegraphics[ width=0.24\textwidth]{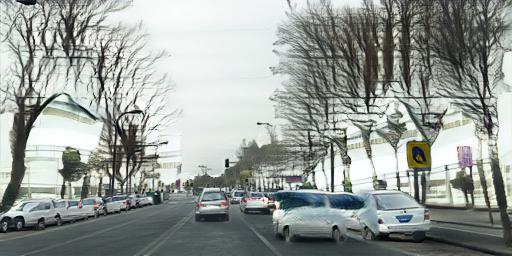}\hspace{-7pt} \\
    \includegraphics[ width=0.24\textwidth]{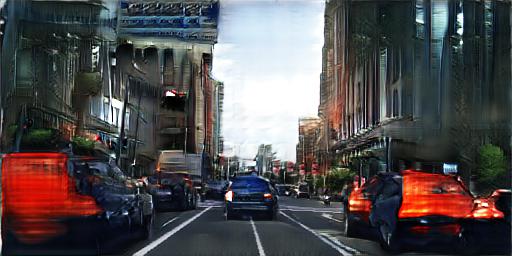}\hspace{-7pt} &
    \includegraphics[ width=0.24\textwidth]{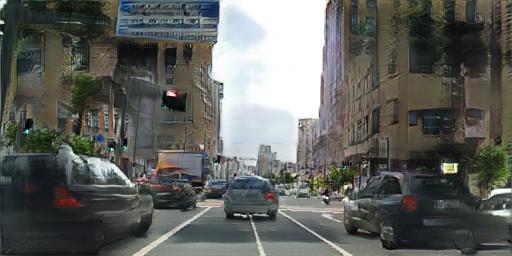}\hspace{-7pt} &
    \includegraphics[ width=0.24\textwidth]{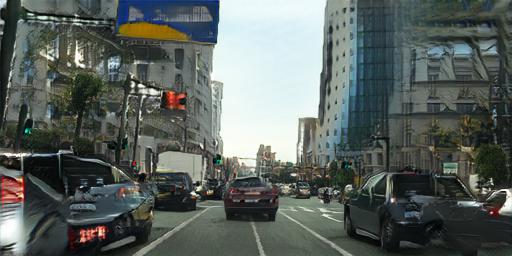}\hspace{-7pt} &
    \includegraphics[ width=0.24\textwidth]{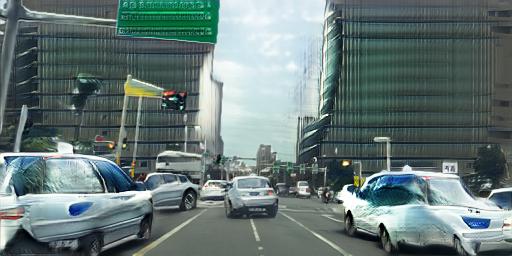}\hspace{-7pt} \\
    \includegraphics[ width=0.24\textwidth]{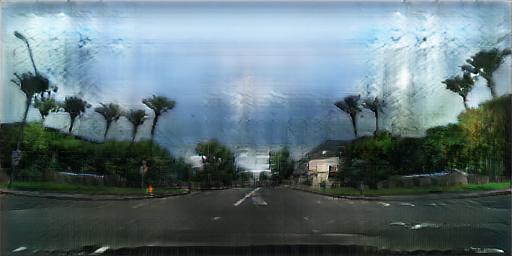}\hspace{-7pt} &
    \includegraphics[ width=0.24\textwidth]{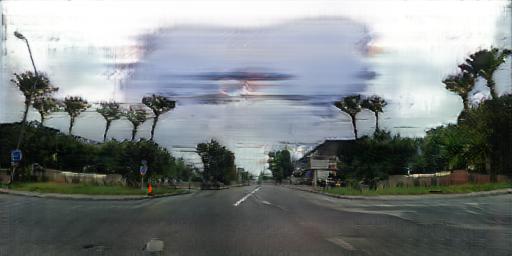}\hspace{-7pt} &
    \includegraphics[ width=0.24\textwidth]{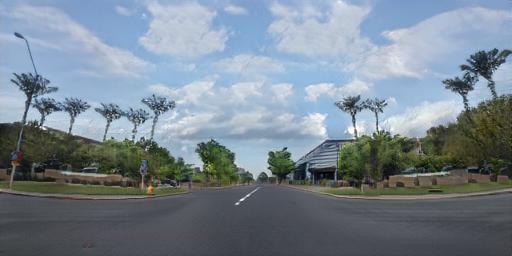}\hspace{-7pt} &
    \includegraphics[ width=0.24\textwidth]{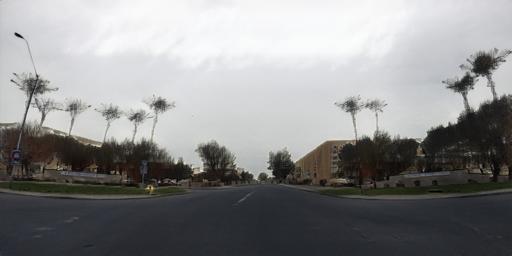}\hspace{-7pt} \\
    \end{tabular}
    \caption{\small\textbf{Qualitative results on Mapillary}. Each row visualizes images synthesized from the same input semantic segmentation mask using different models. All continual models are initialized from the OASIS model that was trained on Cityscapes. The basic cSPADE model, with very small overhead cost, struggles to adapt to the style of Mapillary; indeed we still see the gloomy tone of the previous Cityscapes dataset. Having instance norm helps bring more style transfer effect. Our full CSG0 results look more natural and have less visible artifacts than other models.
    Best viewed in color.}
    \label{fig:qual_res_mapillary}

\end{figure*}

We organize results by the final target dataset in the continual stream, \IE, IDD in Tab.\,\ref{tbl:GC2I} and Fig.\,\ref{fig:qual_res_idd}, Mapillary in Tab.\,\ref{tbl:GC2M} and Fig.\,\ref{fig:qual_res_mapillary}.
In each table, we report results of ablated CSG0 models and the brute-force approach where one full-blown OASIS model is trained for each dataset.

\parag{IDD results.}
With IDD as final target domain, we consider two 2-domain scenarios: \GtoI~(synthetic to real) and \CtoI~(real to real).
From the base OASIS model with $35$ classes, pre-trained either on GTA5 or Cityscapes, CSG0 extends in both cases to $44$ classes of which $9$ are new, only existing in IDD.

The main results are reported in Tab.\,\ref{tbl:GC2I}. In this table, `cSPADE' stands for the basic CSG0 model that adopts the vanilla cSPADE block to accommodate new classes, \EG, only introducing the EL \texttt{conv} layers. Training of the cSPADE model only learns parameters in these EL layers and keeps everything else untouched. 
This basic model, though only introducing a small overhead of 0.3M parameters, does not achieve good results in terms of FID and mIoU; the first column of Figure~\ref{fig:qual_res_idd} visualizes some qualitative results.
We still notice the colors and patterns of the previous domain, \IE, GTA5 in this case; This is most noticeable for  \textit{street} and \textit{vegetable} classes. The synthesized \textit{sky} does not have the cloudy tone as in IDD. For new classes like \textit{tuk-tuk}, the model cannot generate satisfactory results.
When replacing the original BatchNorm layers in cSPADE block with InstanceNorm ones, denoted as `cSPADE$\,{+}\,$IN' in Table~\ref{tbl:GC2I}, the continual model gets better in learning the new style of IDD.
We observe significant improvements in the aforementioned classes.
However, the overall realism level is quite limited, as reflected in the FID score, as well as in the qualitative results shown in column 2 of Figure~\ref{fig:qual_res_idd}.

\NEW{With a larger overhead cost of 10.8M parameters, the final CSG0 model having the weight modulation strategy, denoted as `cSPADE$\,{+}\,$IN$\,{+}\,$WM', obtains much better synthesis quality.
With more learnable parameters, we observe a significant boost as compared to `cSPADE$\,{+}\,$IN'.}
As shown in the third column of Figure~\ref{fig:qual_res_idd}, scenes generated by CSG0 are more realistic with similar color tones as in the IDD dataset.
The \textit{street} areas, sometimes mixing soil and asphalt, look very much like typical Indian roads.

We make some interesting findings when comparing CSG0 with the OASIS model that is trained on IDD,  `OASIS (I)' in Table~\ref{tbl:GC2I}.
The OASIS model obtains a much worse FID of $55.3$ while our CSG0 is able to reach $39.0$ (\GtoI) and $37.0$ (\CtoI).
We posit that such improvements are brought by the  knowledge transferred from the previous domains that our continual models are based on, \IE, GTA5 and Cityscapes.
We note that the OASIS model reported in Table~\ref{tbl:GC2I} corresponds to the brute-force zero-forgetting solution, which resorts to training a full OASIS model to handle a new dataset.

Our CSG0 model, with much smaller overhead cost, outperforms the `OASIS (I)' in terms of FID.
Though on the GAN-test metric, CSG0 models have slightly worse mIoU than the `OASIS (I)', qualitative results of CSG0 are much more convincing.
As visualized in the last column of Figure~\ref{fig:qual_res_idd}, the `OASIS (I)' images have lots of artifacts; most visible are with the trees and the roads.
In general, OASIS images exhibit weird color tone and contrast, making them look less realistic as compared to CSG0's.

\begin{table*}[ht!]
    \setlength\heavyrulewidth{0.25ex}
    \setlength{\tabcolsep}{6pt}
    \aboverulesep=0ex
    \belowrulesep=0ex
    \small
	\centering
	\vspace{-0.2cm}
    \begin{minipage}[c]{0.58\textwidth}
        \begin{center}
        \begin{tabular}{@{}L{0.1cm}@{}@{}L{3.2cm}@{}@{}C{1.2cm}@{}@{}C{1.2cm}@{}@{}C{0.90cm}@{}C{0.90cm}@{}@{}C{0.90cm}@{}C{0.90cm}@{}}
            \multicolumn{8}{c}{(a) Sequences of two datasets}\\
            \toprule
            \multicolumn{2}{c}{\multirow{2}{*}{\small Model}} &  \multirow{2}{*}{\parbox{1.2cm}{\small \centering New\\ params}} &  \multirow{2}{*}{\parbox{1.2cm}{\small \centering Total\\ params}}	& 	\multicolumn{2}{c}{\small G $\rightarrow$ M} &	\multicolumn{2}{c}{\small C $\rightarrow$ M} \\	 
            \cmidrule(lr){5-6}\cmidrule(lr){7-8}
            \multicolumn{1}{c}{}& & & &	FID & mIoU	&	FID & mIoU		\\	
            \midrule
            \multicolumn{1}{l}{\multirow{3}{*}{\rtb{CSG0}}}&\,\,cSPADE & {0.8M} & 72.0M	&	48.6 & 21.6 &	57.6 & 17.6 \\
            &\,\,cSPADE$\,{+}\,$IN	& {0.8M} & 72.0M &	32.6 & 23.3&	38.9 & 21.8 \\
            &\,\,cSPADE$\,{+}\,$IN$\,{+}\,$WM &  11.3M  & 82.5M	&	\textbf{\textcolor{gGreen}{25.1}} & \textbf{\textcolor{gGreen}{26.5}} &	\textbf{\textcolor{gBlue}{24.5}} & \textbf{\textcolor{gBlue}{25.5}} \\
            \midrule
            &\,\,OASIS (M)~\cite{oasis_schonfeld2020you}  & 72M	& 143.2M &	\textcolor{gOrange}{27.1} & \textcolor{gOrange}{30.6}	&	\textit{idem} & \textit{idem}	\\
            \bottomrule
        \end{tabular}
        \vspace{0.2cm}
        \begin{tabular}{@{}L{0.1cm}@{}@{}L{3.2cm}@{}@{}C{1.2cm}@{}@{}C{1.2cm}@{}@{}C{0.90cm}@{}C{0.90cm}@{}@{}C{0.90cm}@{}C{0.90cm}@{}}
            \multicolumn{8}{c}{ }\\
            \multicolumn{8}{c}{(b) Sequences of three datasets}\\
            \toprule
            \multicolumn{2}{c}{\multirow{2}{*}{\small Model}} &  \multirow{2}{*}{\parbox{1.2cm}{\small \centering New\\ params}} &  \multirow{2}{*}{\parbox{1.2cm}{\small \centering Total\\ params}}	& 	\multicolumn{2}{c}{\small G$\rightarrow$I $\rightarrow$ M} &	\multicolumn{2}{c}{\small C$\rightarrow$I $\rightarrow$ M} \\	 
            \cmidrule(lr){5-6}\cmidrule(lr){7-8}
            \multicolumn{1}{c}{}& & & &	FID & mIoU	&	FID & mIoU		\\	
            \midrule
            \multicolumn{1}{l}{\multirow{3}{*}{\rtb{CSG0}}}&\,\,cSPADE & {0.8M} & 82.8M	&	56.2 & 21.6	&	65.8 & 16.9	\\
            &\,\,cSPADE$\,{+}\,$IN & {0.8M}  & 82.8M	&	34.1 & 23.9	&	39.4 & 22.0\\
            &\,\,cSPADE$\,{+}\,$IN$\,{+}\,$WM & 11.3M & 93.3M	&	\textbf{\textcolor{gGreen}{24.0}} & \textbf{\textcolor{gGreen}{26.6}} &	\textbf{\textcolor{gBlue}{25.4}} & \textbf{\textcolor{gBlue}{26.1}}	\\
            \midrule
            &\,\,OASIS (M)~\cite{oasis_schonfeld2020you}  & 72M & 214.6M &	\textcolor{gOrange}{27.1} & \textcolor{gOrange}{30.6} &	\textit{idem} & \textit{idem} \\
            \bottomrule
        \end{tabular}
        \end{center}
	\end{minipage}
	\hfill
    \begin{minipage}[c]{0.4\textwidth}
        \centering
        \includegraphics[width=0.8\textwidth]{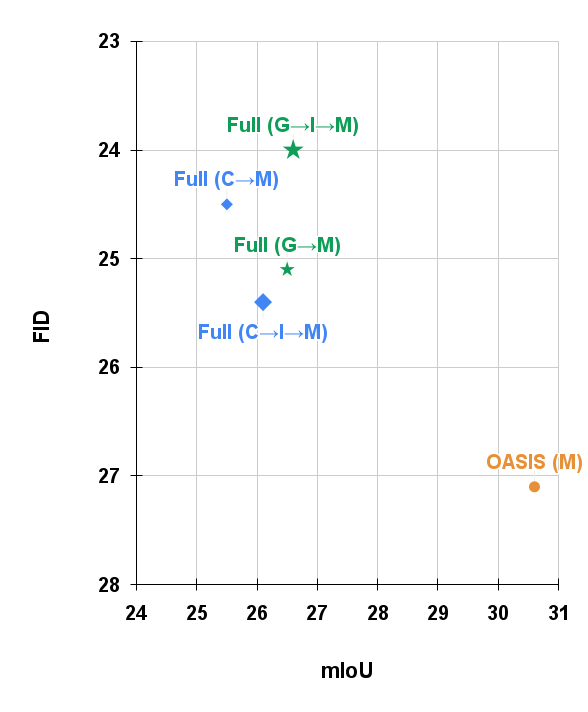}
    \end{minipage}
    \vspace{-0.2cm}
    \caption{\small\textbf{Performance on Mapillary of continual models}. Results of the two continual setups with sequences of (a) two datasets and (b) three datasets are reported. For each sub-table, the structure is the same as in Table~\ref{tbl:GC2I}.
    To the right, we plot the results of our full CSGO models (cSPADE$\,{+}\,$IN$\,{+}\,$WM) in both setups as well as the brute-force model `OASIS (M)'.
    \NEW{CSG0 models outperform OASIS in terms of FID, which shows in the image quality in Figure~\ref{fig:qual_res_mapillary}.}
    Colors are matched between the table and figure.
    }
    \label{tbl:GC2M}
\end{table*}

In addition to the reported metrics, we have tried using generated images for data augmentation.
In detail, we train PSPNet models using both real training IDD data and synthesized data.
Compared to the model trained only on real IDD, which achieves a validation mIoU of $38.7\%$, the model using both real and CSG0 data achieves $39.5\%$ validation mIoU, hence a slight gain of $0.8\%$.

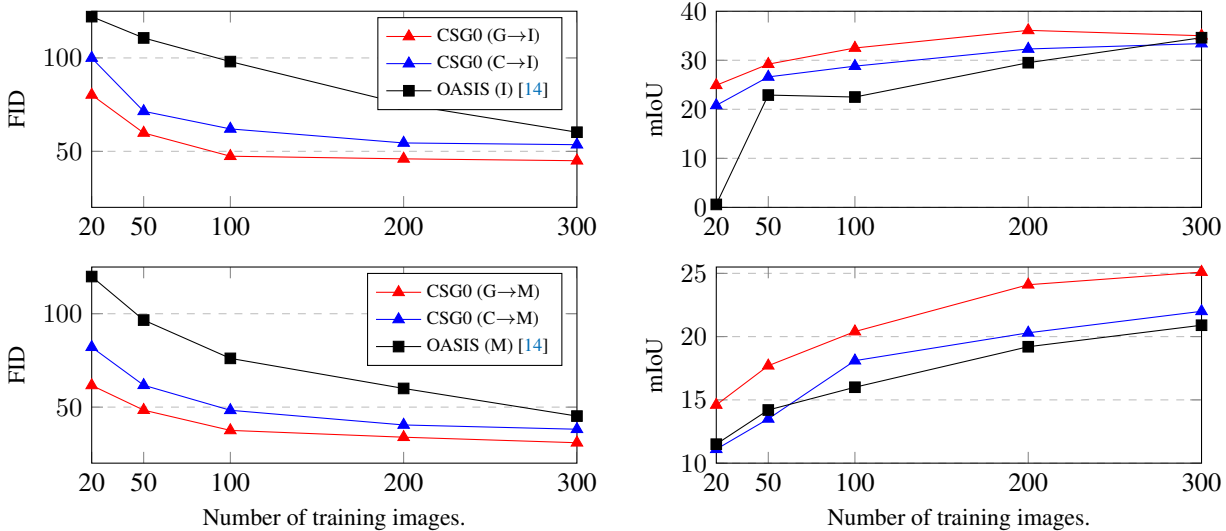
\begin{figure*}[t]
\begin{tabular}{ll}
    \input{plots_arxiv/lowdata_idd_300}&
    \input{plots_arxiv/lowdata_idd_miou_300}\\
\end{tabular}
\begin{tabular}{ll}
    \input{plots_arxiv/lowdata_mapillary_300}&
    \input{plots_arxiv/lowdata_mapillary_miou_300}\\
\end{tabular}
\caption{\small\textbf{Low-data regime training}. We compare our CSG0 \NEW{(`cSPADE$\,{+}\,$IN$\,{+}\,$WM')} models to `OASIS (I)' and `OASIS (M)'. Two sub-figures in the same row share the same legend.
While the OASIS models overfit to the small subsets, thus getting comparatively higher FID, our CSG0 models benefit from transfer learning and achieve better scores. Similar gaps are observed in terms of GAN-test mIoU.}
\label{fig:low_data_regime}
\end{figure*}

\parag{Mapillary results.}
Mapillary has 64 semantic classes in total.
We show results of two continual set-ups with either two or three datasets, respectively in Table~\ref{tbl:GC2M} (a) and (b).
We observe similar results among all CSG0 variants, proving the usefulness of proposed strategies.
The CSG0 models achieve better FID than the brute-force OASIS model.
To learn scene generation for the three datasets, note that our full CSG0 only needs 93.3M parameters in total, less than half of the parameters needed in the brute-force solution.
Comparing results between two and three datasets, we do not see much difference in the real-to-real scenario, \IE, \CtoM~\textit{vs.}~\CtoItoM.
In the synthetic-to-real scenario, with three datasets we notice small FID improvements for some CSG0 models, while GAN-test mIoUs are more or less the same.
Under the same comparison, no significant changes are observed for the brute-force OASIS model.
We conjecture that the number of domains the starting model has been trained on is not a very important factor in zero-forgetting continual scene generation.
In Figure~\ref{fig:qual_res_mapillary}, we visualize some generated Mapillary-like images to demonstrate the differences in synthesis quality.

\begin{figure}[h]
    \hspace{-0.2cm}
    \begin{tabular}{ccc}
    \small CS Masks  & \small OASIS (I)~\cite{oasis_schonfeld2020you} & \small CSG0 \\
    \hspace{-5pt}
    \includegraphics[ width=0.156\textwidth]{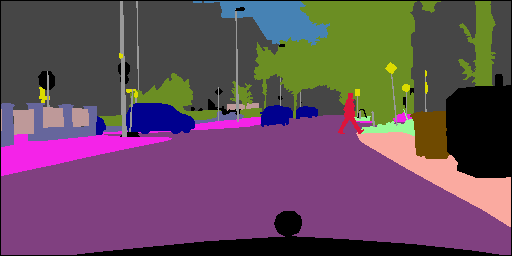} &
    \hspace{-10pt}
    \includegraphics[ width=0.156\textwidth]{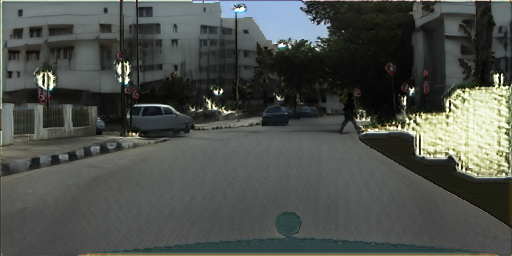} &
    \hspace{-10pt}
    \includegraphics[ width=0.156\textwidth]{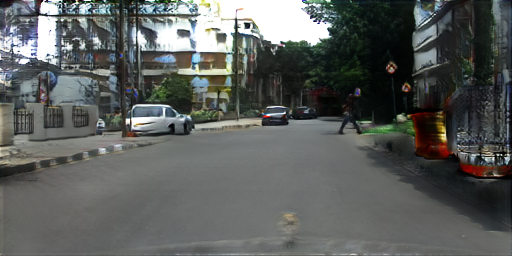} \\
    \hspace{-5pt}
    \includegraphics[ width=0.156\textwidth]{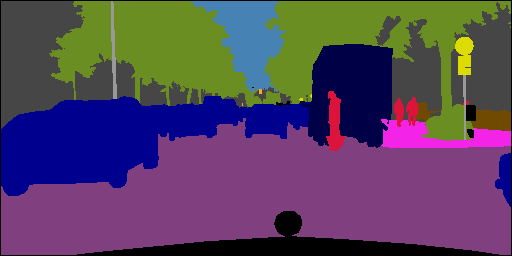} &
    \hspace{-10pt}
    \includegraphics[ width=0.156\textwidth]{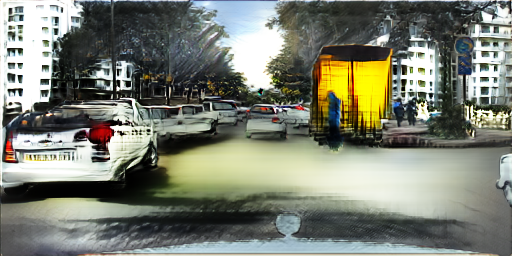} &
    \hspace{-10pt}
    \includegraphics[ width=0.156\textwidth]{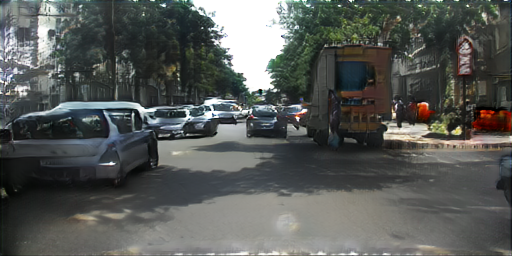}
    \end{tabular}
    \caption{\small \textbf{Cross dataset sampling}. The figure shows images generated from IDD generators conditioned on Cityscapes (CS) masks. 
    `OASIS (I)' does not work well with CS masks thus producing visible artifacts, even on the shared classes.
    }
    \label{fig:cross_dataset_sampling}
\end{figure}

\parag{Advantage in cross-dataset sampling.}
One interesting property of CSG0 is its advantage in cross-dataset sampling, thanks to continual learning.
In Figure~\ref{fig:cross_dataset_sampling}, we illustrate some results in which we sample images having Cityscapes layouts but with IDD style.
That is made possible by feeding Cityscapes segmentation masks into models trained on IDD.
The task is challenging: If the GAN models cannot handle well the shift in semantic distributions between Cityscapes and IDD, the synthesis quality would be greatly degraded.
Compared to `OASIS (I)', CSG0 produces more realistic results with fewer artifacts.

\subsection{Low-data regime}
Not only memory efficient, our continual framework allows seamless knowledge transfer from the previous dataset, encoded in the learned parameters, to a new yet related dataset.
In this experiment, we showcase the merit of CSG0 when training with limited supervision, \IE, the low-data regime, in the target dataset.
In detail, we trained CSG0 models using IDD and Mapillary subsets of $20$, $50$, $100$, $200$ and $300$ data samples respectively.
Our models are initialized using the OASIS model pre-trained on either GTA5 or Cityscapes.
We compare against the full OASIS model trained only on similar subsets.

Figure~\ref{fig:low_data_regime} plots performance curves of different models.
CSG0 outperforms the OASIS models by significant margin, especially in the extreme set-ups with very little supervision, \EG, only 20 and 50 training samples.
Having more training data further closes the performance gap between CSG0 and OASIS.
Results in the low-data regime confirm the benefit of transfer learning in our continual framework.

%% file: plots_arxiv/lowdata_idd_300.tex
\begin{tikzpicture}
\begin{axis}[
    legend style={font=\scriptsize},
    legend cell align={left},
    width=0.46\textwidth, height=0.24\textwidth,
    ylabel={\small{FID}},
    xmin=20, xmax=300,
    ymin=20, ymax=125,
    xtick={20, 50, 100, 200, 300},
    xticklabels={20, 50, 100, 200, 300},
    xticklabel style={rotate=0},
    legend pos = north east,
    ymajorgrids=true,
    grid style=dashed,]
\addplot[red, mark=triangle*, mark size=2.5pt] table [x=size, y=fid_gi, col sep=space] {plots_arxiv/lowdata_results.csv};
\addplot[blue, mark=triangle*, mark size=2.5pt] table [x=size, y=fid_ci, col sep=space] {plots_arxiv/lowdata_results.csv};
\addplot[black, mark=square*, mark size=2pt] table [x=size, y=oasis_i, col sep=space] {plots_arxiv/lowdata_results.csv};
\legend{CSG0 (G$\rightarrow$I), CSG0 (C$\rightarrow$I), OASIS (I)~\cite{oasis_schonfeld2020you}}
\end{axis}
\end{tikzpicture}

%% file: plots_arxiv/lowdata_idd_miou_300.tex
\begin{tikzpicture}
\begin{axis}[
    legend style={font=\scriptsize},
    legend cell align={left},
    width=0.46\textwidth, height=0.24\textwidth,
    ylabel={\small{mIoU}},
    xmin=20, xmax=300,
    ymin=0, ymax=40,
    xtick={20, 50, 100, 200, 300},
    xticklabels={20, 50, 100, 200, 300},
    legend pos = south east,
    ymajorgrids=true,
    grid style=dashed,]
\addplot[red, mark=triangle*, mark size=2.5pt] table [x=size, y=g2i, col sep=space] {plots_arxiv/lowdata_results_miou.csv};
\addplot[blue, mark=triangle*, mark size=2.5pt] table [x=size, y=c2i, col sep=space] {plots_arxiv/lowdata_results_miou.csv};
\addplot[black, mark=square*, mark size=2pt] table [x=size, y=oasis_i, col sep=space] {plots_arxiv/lowdata_results_miou.csv};;
\end{axis}
\end{tikzpicture}

%% file: plots_arxiv/lowdata_mapillary_300.tex
\begin{tikzpicture}
\begin{axis}[
    legend style={font=\scriptsize},
    legend cell align={left},
    width=0.46\textwidth, height=0.24\textwidth,
    xlabel={\small{Number of training images.}},
    ylabel={\small{FID}},
    xmin=20, xmax=300,
    ymin=20, ymax=125,
    xtick={20, 50, 100, 200, 300},
    xticklabels={20, 50, 100, 200, 300},
    xticklabel style={rotate=0},
    legend pos = north east,
    ymajorgrids=true,
    grid style=dashed,]
\addplot[red, mark=triangle*, mark size=2.5pt] table [x=size, y=fid_gm, col sep=space] {plots_arxiv/lowdata_results.csv};
\addplot[blue, mark=triangle*, mark size=2.5pt] table [x=size, y=fid_cm, col sep=space] {plots_arxiv/lowdata_results.csv};
\addplot[black, mark=square*, mark size=2pt] table [x=size, y=oasis_m, col sep=space] {plots_arxiv/lowdata_results.csv};

\legend{CSG0 (G$\rightarrow$M), CSG0 (C$\rightarrow$M), OASIS (M)~\cite{oasis_schonfeld2020you}}
\end{axis}
\end{tikzpicture}

%% file: plots_arxiv/lowdata_mapillary_miou_300.tex
\begin{tikzpicture}
\begin{axis}[
    legend style={font=\scriptsize},
    legend cell align={left},
    width=0.46\textwidth, height=0.24\textwidth,
    xlabel={\small{Number of training images.}},
    ylabel={\small{mIoU}},
    xmin=20, xmax=300,
    ymin=10, ymax=25.5,
    xtick={20, 50, 100, 200, 300},
    xticklabels={20, 50, 100, 200, 300},
    legend pos = south east,
    ymajorgrids=true,
    grid style=dashed,]
\addplot[red, mark=triangle*, mark size=2.5pt] table [x=size, y=g2m, col sep=space] {plots_arxiv/lowdata_results_miou.csv};
\addplot[blue, mark=triangle*, mark size=2.5pt] table [x=size, y=c2m, col sep=space] {plots_arxiv/lowdata_results_miou.csv};
\addplot[black, mark=square*, mark size=2pt] table [x=size, y=oasis_m, col sep=space] {plots_arxiv/lowdata_results_miou.csv};
\end{axis}
\end{tikzpicture}

%% file: CLVISION_conclusion.tex
This work addresses a pragmatic task of continual semantic scene generation with zero-forgetting. 
We propose a modular framework, named CSG0, with novel architecture designs and strategies for this task.
To showcase the merit of our framework, we conduct intensive experiments on various continual urban scene setups, covering both synthetic-to-real and real-to-real scenarios. Quantitative evaluations and qualitative visualizations demonstrate the interest of our  CSG0 framework, which operates with minimal overhead cost (in terms of architecture size and training).
Benefiting from continual learning, CSG0 outperforms the state-of-the-art OASIS model trained on single domains. We also provide experiments with three datasets to emphasize how well our strategy generalizes despite its cost constraints. 

%% file: CLVISION_appendix.tex
\begin{figure*}
    \hspace{-0.2cm}
    \small
    \begin{tabular}{cccc}
    cSPADE&cSPADE$\,{+}\,$IN&cSPADE$\,{+}\,$IN$\,{+}\,$WM& OASIS(IDD)\\
    \includegraphics[ width=0.24\textwidth]{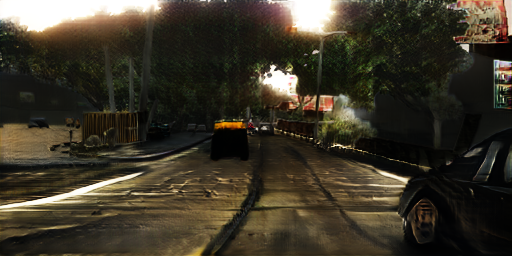}\hspace{-7pt} &
    \includegraphics[ width=0.24\textwidth]{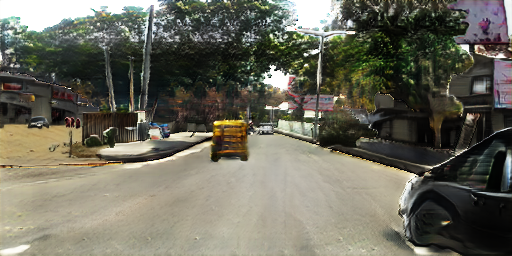}\hspace{-7pt} &
    \includegraphics[ width=0.24\textwidth]{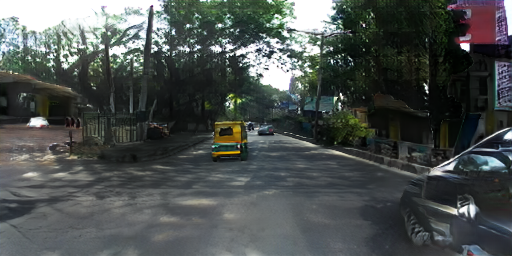}\hspace{-7pt} &
    \includegraphics[ width=0.24\textwidth]{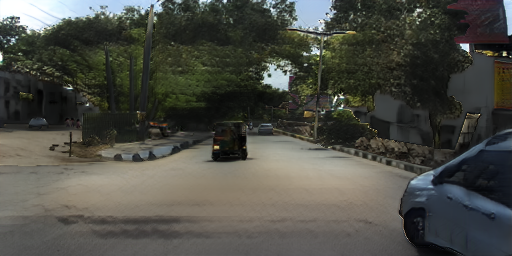}\hspace{-7pt} \\
    
    \includegraphics[ width=0.24\textwidth]{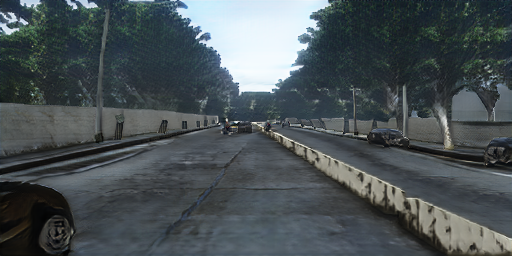}\hspace{-7pt} &
    \includegraphics[ width=0.24\textwidth]{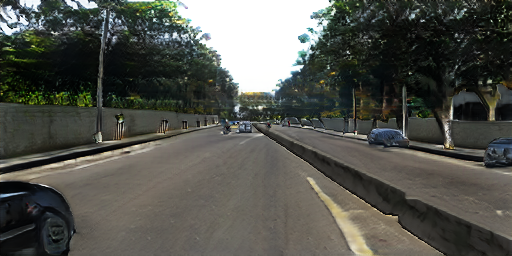}\hspace{-7pt} &
    \includegraphics[ width=0.24\textwidth]{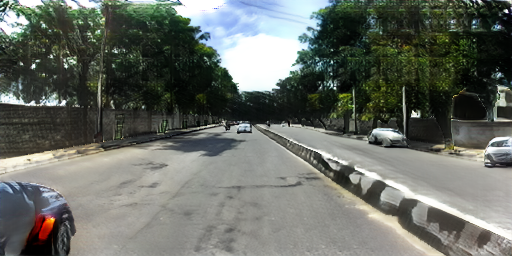}\hspace{-7pt} &
    \includegraphics[ width=0.24\textwidth]{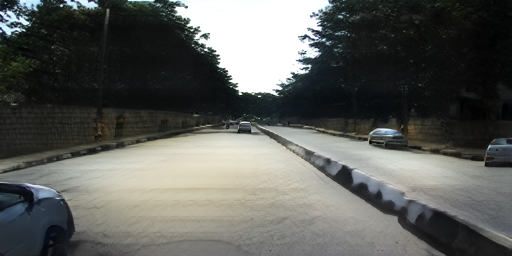}\hspace{-7pt} \\
    
    \includegraphics[ width=0.24\textwidth]{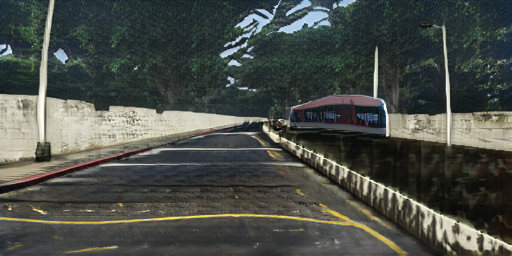}\hspace{-7pt} &
    \includegraphics[ width=0.24\textwidth]{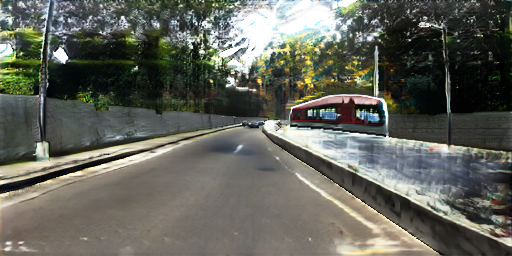}\hspace{-7pt} &
    \includegraphics[ width=0.24\textwidth]{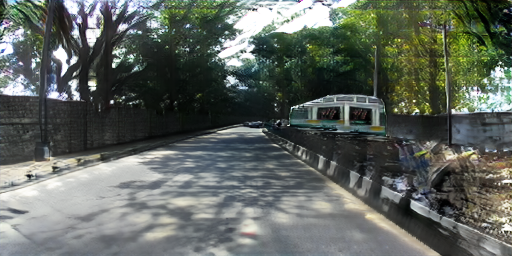}\hspace{-7pt} &
    \includegraphics[ width=0.24\textwidth]{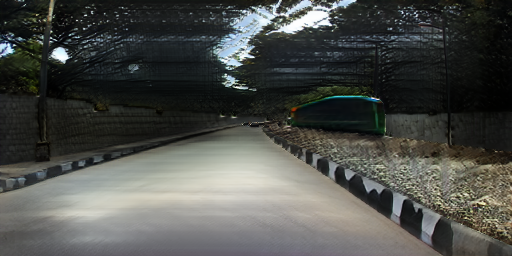}\hspace{-7pt} \\
    
    \includegraphics[ width=0.24\textwidth]{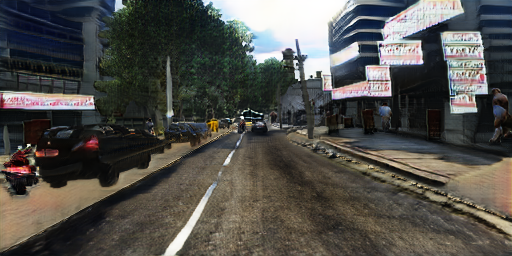}\hspace{-7pt} &
    \includegraphics[ width=0.24\textwidth]{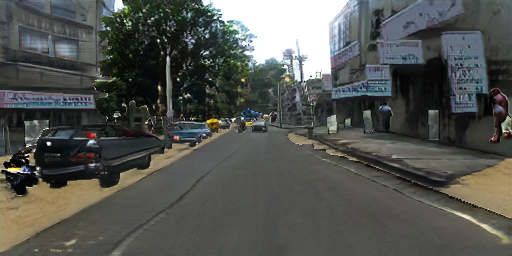}\hspace{-7pt} &
    \includegraphics[ width=0.24\textwidth]{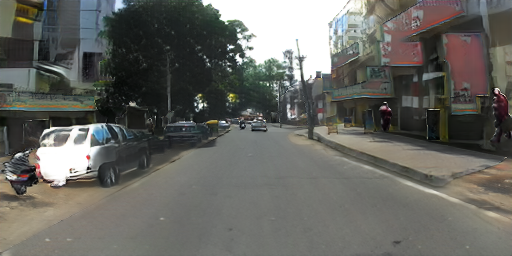}\hspace{-7pt} &
    \includegraphics[ width=0.24\textwidth]{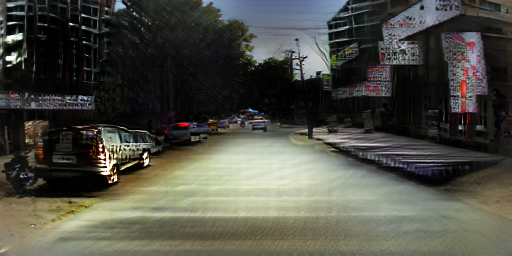}\hspace{-7pt} \\
    
    \includegraphics[ width=0.24\textwidth]{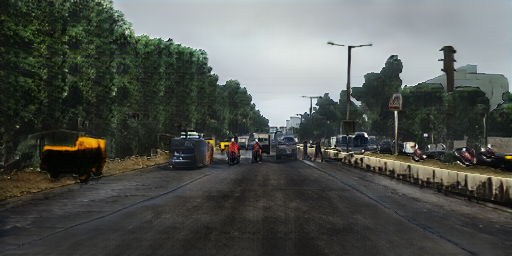}\hspace{-7pt} &
    \includegraphics[ width=0.24\textwidth]{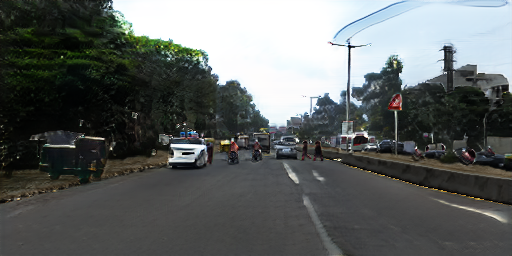}\hspace{-7pt} &
    \includegraphics[ width=0.24\textwidth]{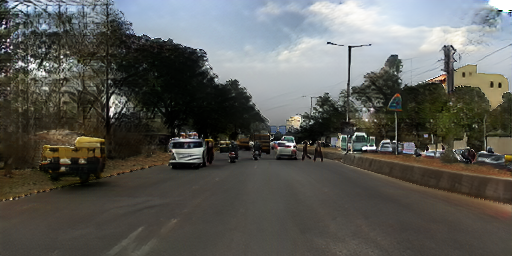}\hspace{-7pt} &
    
    \includegraphics[ width=0.24\textwidth]{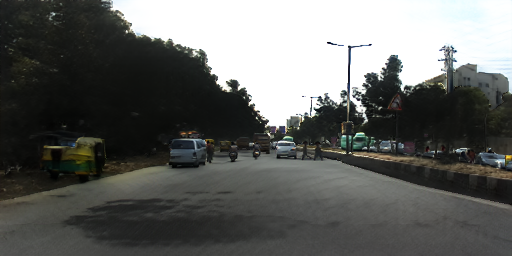}\hspace{-7pt} \\
    
    \includegraphics[ width=0.24\textwidth]{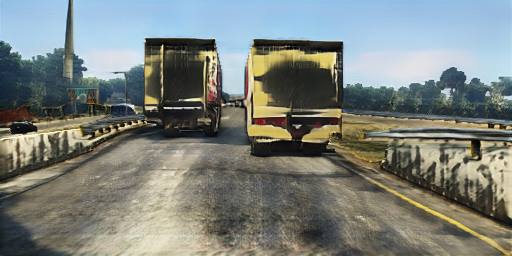}\hspace{-7pt} &
    \includegraphics[ width=0.24\textwidth]{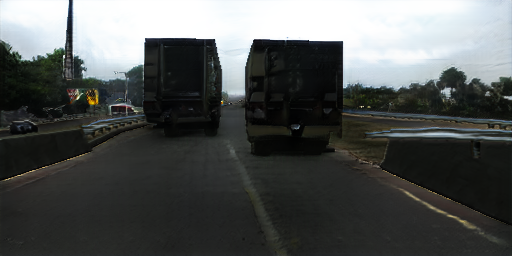}\hspace{-7pt} &
    \includegraphics[ width=0.24\textwidth]{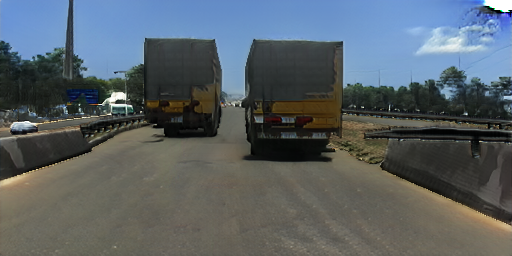}\hspace{-7pt} &
    \includegraphics[ width=0.24\textwidth]{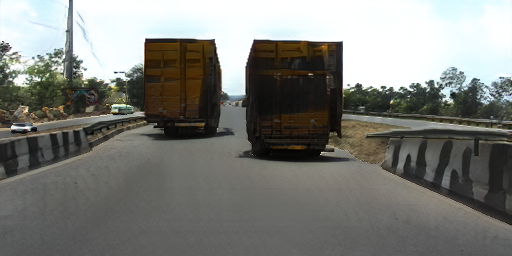}\hspace{-7pt} \\
    
    \includegraphics[ width=0.24\textwidth]{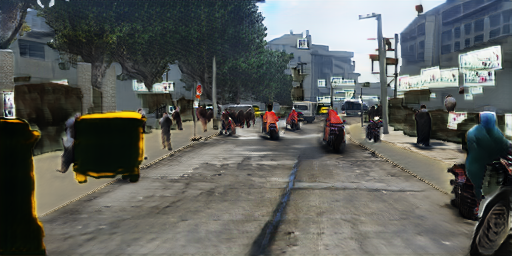}\hspace{-7pt} &
    \includegraphics[ width=0.24\textwidth]{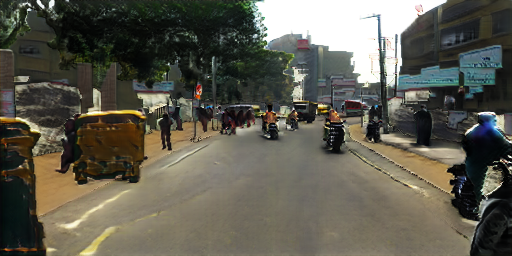}\hspace{-7pt} &
    \includegraphics[ width=0.24\textwidth]{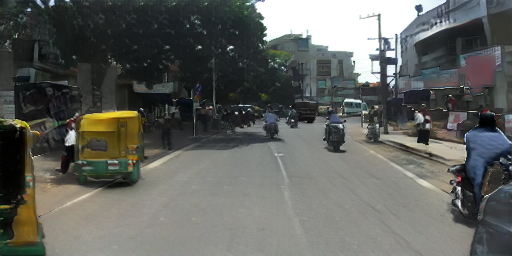}\hspace{-7pt} &
    \includegraphics[ width=0.24\textwidth]{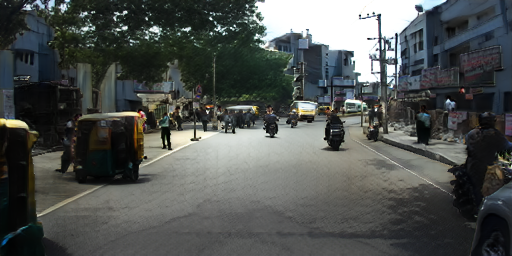}\hspace{-7pt} \\
    
    \includegraphics[ width=0.24\textwidth]{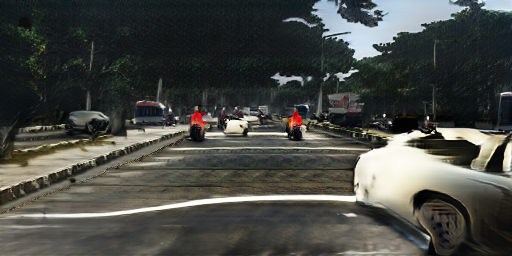}\hspace{-7pt} &
    \includegraphics[ width=0.24\textwidth]{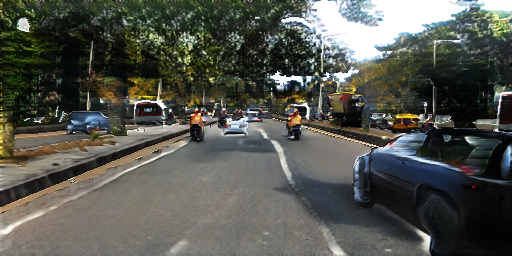}\hspace{-7pt} &
    \includegraphics[ width=0.24\textwidth]{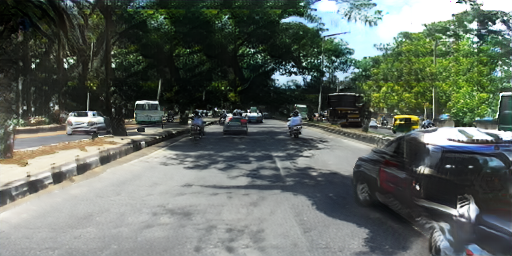}\hspace{-7pt} &
    \includegraphics[ width=0.24\textwidth]{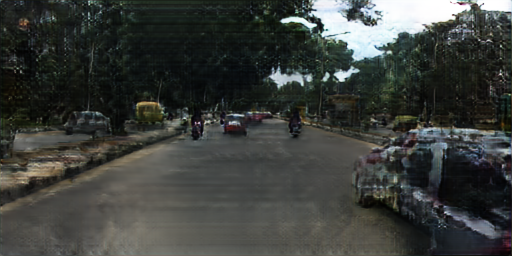}\hspace{-7pt} \\
    
    \includegraphics[ width=0.24\textwidth]{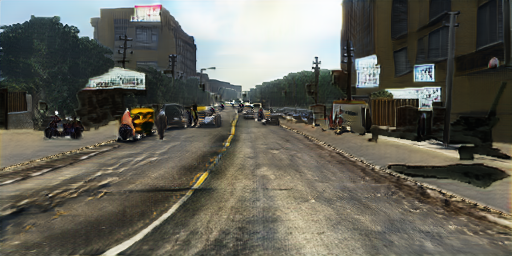}\hspace{-7pt} &
    \includegraphics[ width=0.24\textwidth]{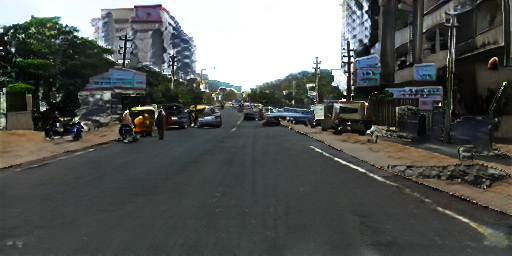}\hspace{-7pt} &
    \includegraphics[ width=0.24\textwidth]{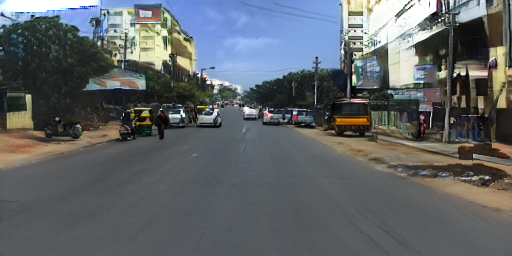}\hspace{-7pt} &
    \includegraphics[ width=0.24\textwidth]{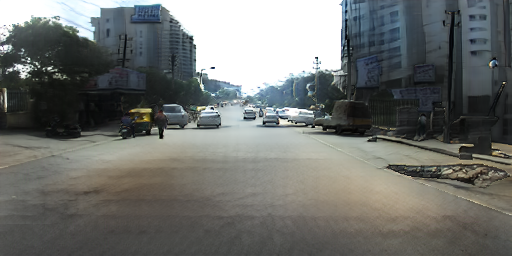}\hspace{-7pt} \\
    
    \end{tabular}
    \vspace{-0.2cm}
    \caption{\small\textbf{Qualitative results on IDD} -- extending Figure~\ref{fig:qual_res_idd}. Each row visualizes images synthesized from the same input semantic segmentation mask using different models. All continual models are initialized from the OASIS model that was trained on GTA5. 
    Best viewed in color.}
    \label{fig:qual_res_idd_suppmat}
    \vspace{-0.1cm}
\end{figure*}

\clearpage
\begin{figure*}
    \hspace{-0.2cm}
    \small
    \begin{tabular}{cccc}
    cSPADE&cSPADE$\,{+}\,$IN&cSPADE$\,{+}\,$IN$\,{+}\,$WM& OASIS(Mapillary)\\
    
    \includegraphics[ width=0.24\textwidth]{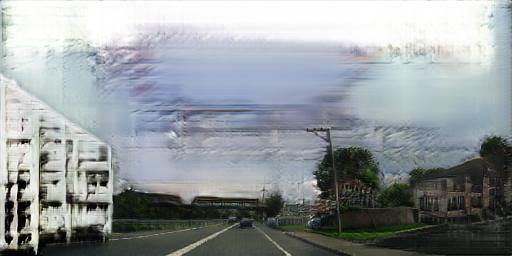}\hspace{-7pt} &
    \includegraphics[ width=0.24\textwidth]{images_arxiv/qual/cityscapes2mapillary_lll_instance_sum_oldbase/_EGt7GTUW6OKw0zzz7c5-A.jpg}\hspace{-7pt} &
    \includegraphics[ width=0.24\textwidth]{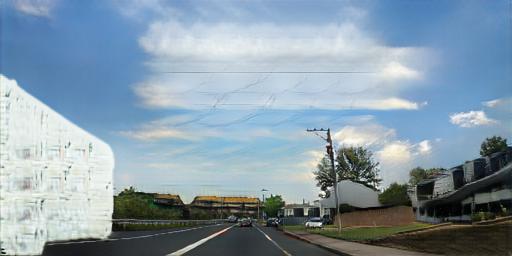}\hspace{-7pt} &
    \includegraphics[ width=0.24\textwidth]{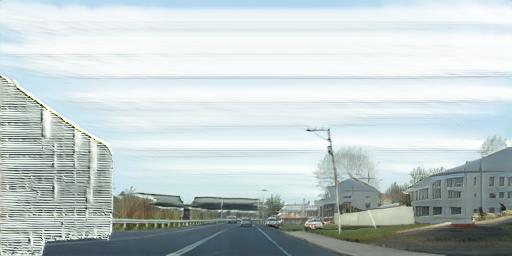}\hspace{-7pt} \\

    \includegraphics[ width=0.24\textwidth]{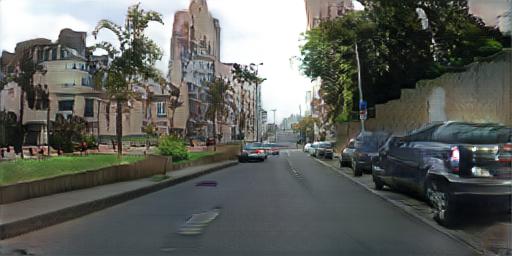}\hspace{-7pt} &
    \includegraphics[ width=0.24\textwidth]{images_arxiv/qual/cityscapes2mapillary_lll_instance_sum_oldbase/_Q-pepPGJ5BackpdSQclrg.jpg}\hspace{-7pt} &
    \includegraphics[ width=0.24\textwidth]{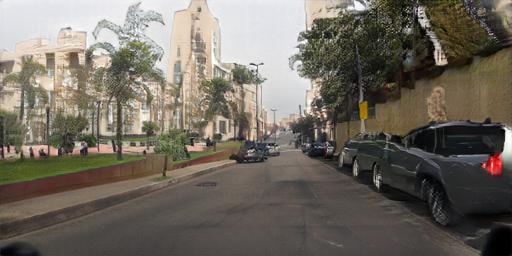}\hspace{-7pt} &
    \includegraphics[ width=0.24\textwidth]{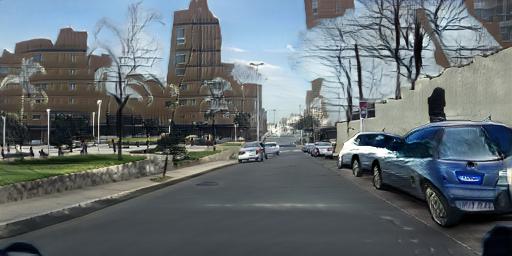}\hspace{-7pt} \\
    
    \includegraphics[ width=0.24\textwidth]{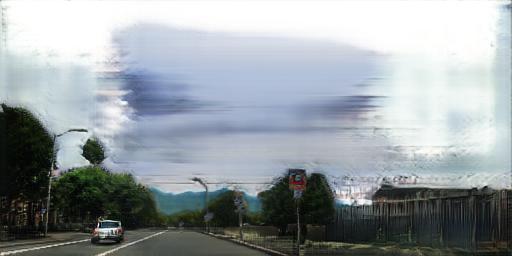}\hspace{-7pt} &
    \includegraphics[ width=0.24\textwidth]{images_arxiv/qual/cityscapes2mapillary_lll_instance_sum_oldbase/_QDf_pTzs-sEyBubXqmqxA.jpg}\hspace{-7pt} &
    \includegraphics[ width=0.24\textwidth]{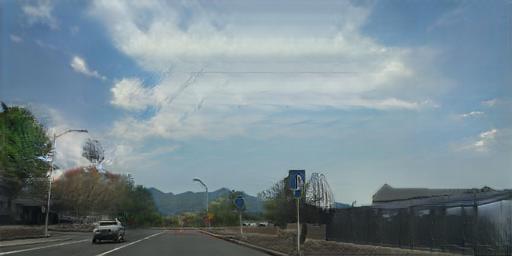}\hspace{-7pt} &
    \includegraphics[ width=0.24\textwidth]{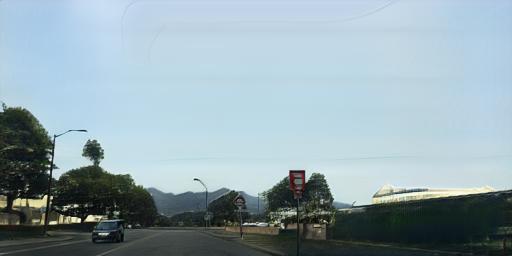}\hspace{-7pt} \\
    
    \includegraphics[ width=0.24\textwidth]{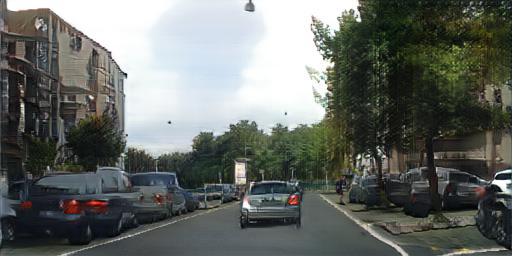}\hspace{-7pt} &
    \includegraphics[ width=0.24\textwidth]{images_arxiv/qual/cityscapes2mapillary_lll_instance_sum_oldbase/61pk4tz3PHmfDkc8VBBtsg.jpg}\hspace{-7pt} &
    \includegraphics[ width=0.24\textwidth]{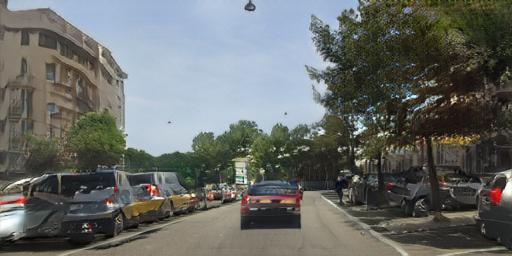}\hspace{-7pt} &
    \includegraphics[ width=0.24\textwidth]{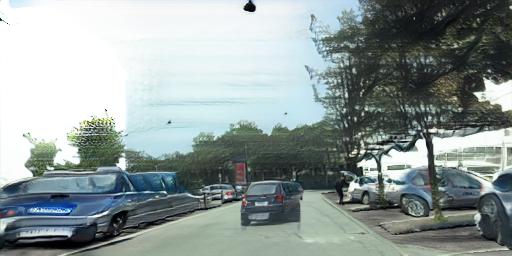}\hspace{-7pt} \\
    
    \includegraphics[ width=0.24\textwidth]{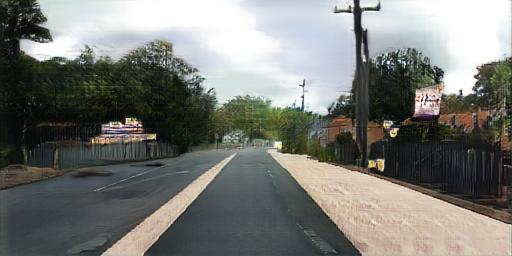}\hspace{-7pt} &
    \includegraphics[ width=0.24\textwidth]{images_arxiv/qual/cityscapes2mapillary_lll_instance_sum_oldbase/Q26wimQNy9NCcUCPNiQ5Kw.jpg}\hspace{-7pt} &
    \includegraphics[ width=0.24\textwidth]{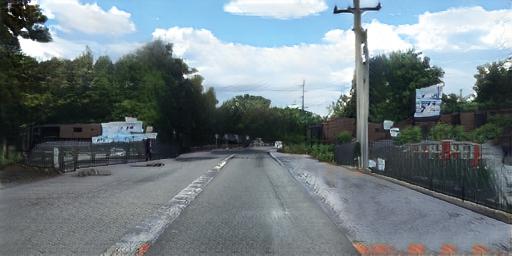}\hspace{-7pt} &
    \includegraphics[ width=0.24\textwidth]{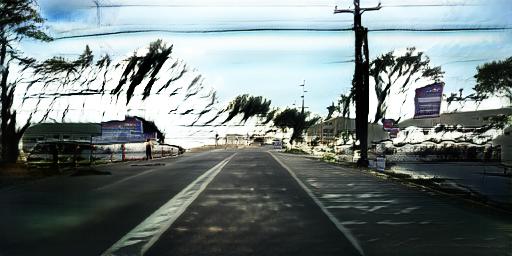}\hspace{-7pt} \\
    
    \includegraphics[ width=0.24\textwidth]{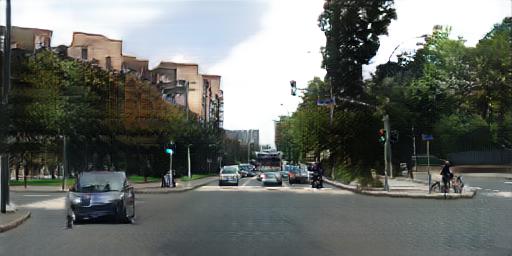}\hspace{-7pt} &
    \includegraphics[ width=0.24\textwidth]{images_arxiv/qual/cityscapes2mapillary_lll_instance_sum_oldbase/Tc1mSICkW-NQbcUKisACQw.jpg}\hspace{-7pt} &
    \includegraphics[ width=0.24\textwidth]{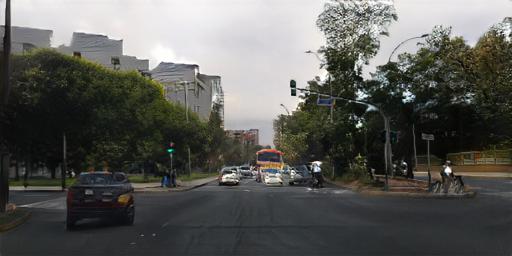}\hspace{-7pt} &
    \includegraphics[ width=0.24\textwidth]{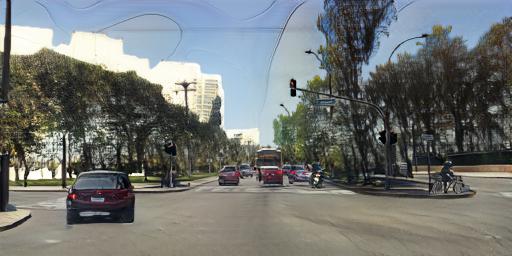}\hspace{-7pt} \\

    \includegraphics[ width=0.24\textwidth]{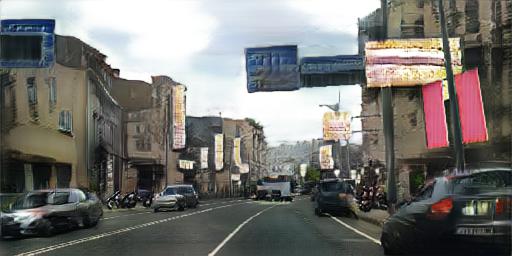}\hspace{-7pt} &
    \includegraphics[ width=0.24\textwidth]{images_arxiv/qual/cityscapes2mapillary_lll_instance_sum_oldbase/_G--4T8xsKtSDNBXVPDrxg.jpg}\hspace{-7pt} &
    \includegraphics[ width=0.24\textwidth]{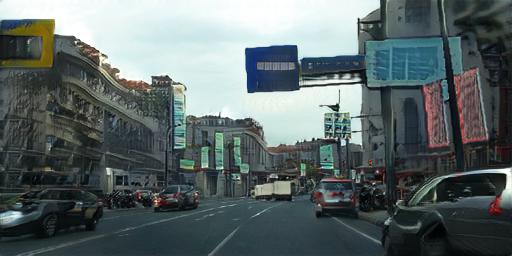}\hspace{-7pt} &
    \includegraphics[ width=0.24\textwidth]{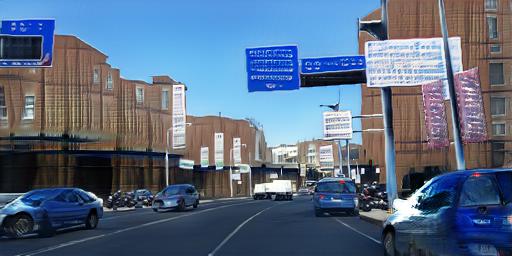}\hspace{-7pt} \\
    
    \includegraphics[ width=0.24\textwidth]{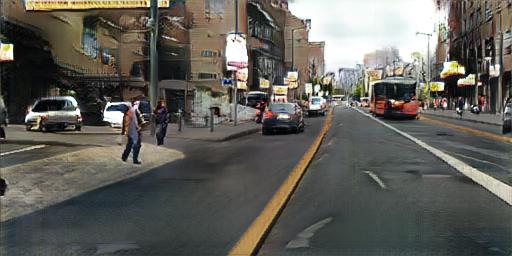}\hspace{-7pt} &
    \includegraphics[ width=0.24\textwidth]{images_arxiv/qual/cityscapes2mapillary_lll_instance_sum_oldbase/_hEblyn68AdOapTDv7doCg.jpg}\hspace{-7pt} &
    \includegraphics[ width=0.24\textwidth]{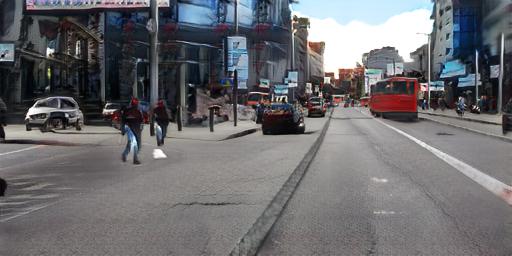}\hspace{-7pt} &
    \includegraphics[ width=0.24\textwidth]{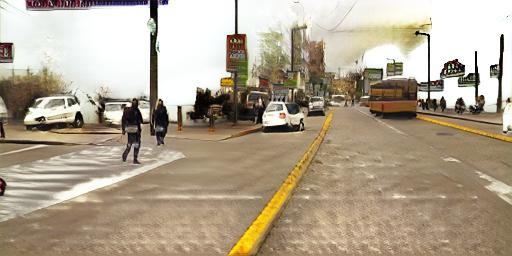}\hspace{-7pt} \\
    
    \includegraphics[ width=0.24\textwidth]{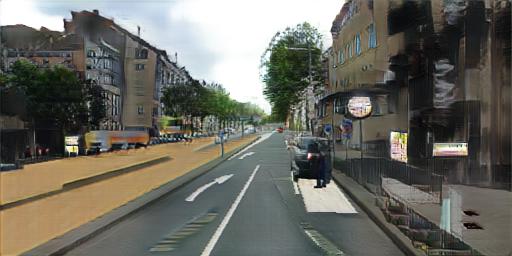}\hspace{-7pt} &
    \includegraphics[ width=0.24\textwidth]{images_arxiv/qual/cityscapes2mapillary_lll_instance_sum_oldbase/0-g5x1x9t7t6_lmXUPFazw.jpg}\hspace{-7pt} &
    \includegraphics[ width=0.24\textwidth]{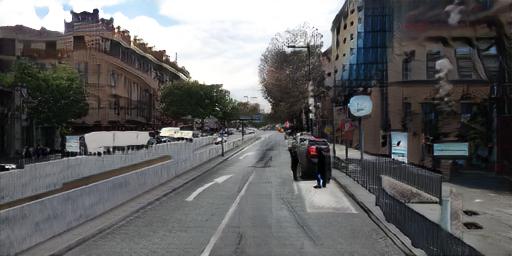}\hspace{-7pt} &
    \includegraphics[ width=0.24\textwidth]{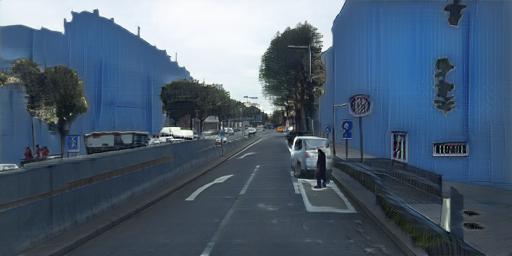}\hspace{-7pt} \\
    
    \end{tabular}
    \vspace{-0.2cm}
    \caption{\small\textbf{Qualitative results on Mapillary} -- extending Figure~\ref{fig:qual_res_mapillary}. Each row visualizes images synthesized from the same input semantic segmentation mask using different models. All continual models are initialized from the OASIS model that was trained on Cityscapes.
    Best viewed in color.}
    \label{fig:qual_res_mapillary_suppmat}
    \vspace{-0.4cm}
\end{figure*}

\clearpage
\begin{table*}[h!]
    \setlength\heavyrulewidth{0.25ex}
    \aboverulesep=0ex
    \belowrulesep=0ex
    \scriptsize
    \caption{\small \textbf{Class mappings in continual urban scene generation.} `Orig. ID' and `Cont. Id' are the original class indices and the corresponding indices used in continual setups. (a) Semantic classes of GTA5/Cityscapes. (b) The 44 used IDD classes: the classes that extend the GTA5/Cityscapes label space are highlighted in \textcolor{cidd}{magenta}. (c) The 64 used Mapillary classes: extended Mapillary classes are highlighted in
    \textcolor{cmapillary}{green}. \textcolor{cnot_idd}{Gray} classes in (b) and (c) are not present in IDD and Mapillary respectively. We redefine some of the Mapillary classes as the union of a few original Mapillary classes; these can be noted by repetition of class ids in `Cont. Id'.}
    \vspace{-0.28cm}
    \begin{minipage}[t]{0.2\textwidth}
        \vspace{0pt}
        \begin{tabu}{rcc}
            \toprule
            Name & Orig. Id & Cont. Id\\
            \midrule
            unlabeled & 0 & 0\\
            ego vehicle & 1&1\\
            rectification border & 2&2\\
            out of roi & 3&3\\
            static & 4&4\\
            dynamic & 5&5\\
            ground & 6&6\\
            road & 7& 7\\
            sidewalk & 8&8\\
            parking & 9 &9\\
            rail track & 10&10\\
            building & 11&11\\
            wall & 12&12\\
            fence & 13&13\\
            guard rail & 14&14\\
            bridge & 15&15\\
            tunnel & 16&16\\
            pole & 17&17\\
            polegroup & 18&18\\
            traffic light & 19&19\\
            traffic sign & 20&20\\
            vegetation & 21&21\\
            terrain & 22&22\\
            sky & 23&23\\
            person & 24&24\\
            rider & 25&25\\
            car & 26&26\\
            truck & 27&27\\
            bus & 28&28\\
            caravan & 29&29\\
            trailer & 30&30\\
            train & 31&31\\
            motorcycle & 32&32\\
            bicycle & 33&33\\
            license plate & -1&34\\
            \bottomrule
            \multicolumn{3}{c}{\scriptsize \textbf{ (a) GTA5 and Cityscapes semantic classes.}}
        \end{tabu}
        \vspace{0pt}
    \end{minipage}
    \hspace{2.2cm}
    \begin{minipage}[t]{0.2\textwidth}
        \vspace{0pt}
        \begin{tabu}{rcc}
            \toprule
            Name & Orig. Id & Cont. Id\\
            \midrule
            road & 0 & 7\\
            parking & 1 & 9\\
            \rowfont{\color{cidd}}drivable fallback & 2 & 35\\
            sidewalk & 3 & 8\\
            rail track & 4 & 10\\
            \rowfont{\color{cidd}}non-drivable fallback & 5 & 36\\
            person & 6 & 24\\
            \rowfont{\color{cidd}}animal & 7 & 37\\
            rider & 8 & 25\\
            motorcycle & 9 & 32\\
            bicycle & 10 & 33\\
            \rowfont{\color{cidd}}autorickshaw & 11 & 38\\
            car & 12 & 26\\
            truck & 13 & 27\\
            bus & 14 & 28\\
            caravan & 15 & 29\\
            trailer & 16 & 30\\
            train & 17 & 31\\
            \rowfont{\color{cidd}}vehicle fallback & 18 & 39\\
            \rowfont{\color{cidd}}curb & 19 & 40\\
            wall & 20 & 12\\
            fence & 21 & 13\\
            guard rail & 22 & 14\\
            \rowfont{\color{cidd}}billboard & 23 & 41\\
            traffic sign & 24 & 20\\
            traffic light & 25 & 19\\
            pole & 26 & 17\\
            polegroup & 27 & 18\\
            \rowfont{\color{cidd}}obs-str-bar-fallback & 28 & 42\\
            building & 29 & 11\\
            bridge & 30 & 15\\
            tunnel & 31 & 16\\
            vegetation & 32 & 21\\
            sky & 33 & 23\\
            \rowfont{\color{cidd}}fallback background & 34 & 43\\
            unlabeled & 35 & 0\\
            ego vehicle & 36 & 1\\
            rectification border & 37 & 2\\
            out of roi & 38 & 3\\
            license plate & 39 & 34\\
            \midrule
            \rowfont{\color{cnot_idd}}static&-1&4\\
            \rowfont{\color{cnot_idd}}dynamic&-1&5\\
            \rowfont{\color{cnot_idd}}ground&-1&6\\
            \rowfont{\color{cnot_idd}}terrain&-1&22\\
            \bottomrule
            \multicolumn{3}{c}{\scriptsize \textbf{ (b) IDD semantic classes.}}
        \end{tabu}
        \vspace{0pt}
    \end{minipage}
    \hspace{2.2cm}
    \begin{minipage}[t]{0.2\textwidth}
        \vspace{0pt}
        \begin{tabu}{rcc}
            \toprule
            Name & Orig. Id & Cont. Id \\
            \midrule
            \rowfont{\color{cmapillary}}bird & 0 & 35\\
            \rowfont{\color{cmapillary}}ground animal & 1 & 36\\
            \rowfont{\color{cmapillary}}curb & 2 & 37\\
            fence & 3 & 13\\
            guard rail & 4 & 14\\
            \rowfont{\color{cmapillary}}barrier & 5 & 38\\
            wall & 6 & 12\\
            \rowfont{\color{cmapillary}}bike lane & 7 &  39\\
            \rowfont{\color{cmapillary}}crosswalk - plain & 8 & 40\\
            \rowfont{\color{cmapillary}}curb cut & 9 & 37\\
            parking & 10 & 9\\
            \rowfont{\color{cmapillary}}pedestrian area & 11 & 41\\
            rail track & 12 & 10\\
            road & 13 & 7\\
            \rowfont{\color{cmapillary}}service lane & 14 & 39\\
            sidewalk & 15 & 8\\
            bridge & 16 & 15\\
            building & 17 & 11\\
            tunnel & 18 & 16\\
            person & 19 & 24\\
            bicyclist & 20 & 25\\
            motorcyclist & 21 & 25\\
            other rider & 22 & 25\\
            \rowfont{\color{cmapillary}}lane marking - crosswalk & 23 & 40\\
            \rowfont{\color{cmapillary}}lane marking - general  & 24 & 39\\
            \rowfont{\color{cmapillary}}mountain & 25 & 42\\
            \rowfont{\color{cmapillary}}sand & 26 & 43\\
            sky & 27 & 23\\
            \rowfont{\color{cmapillary}}snow & 28 & 44\\
            terrain & 29 & 22\\
            vegetation & 30 & 21\\
            \rowfont{\color{cmapillary}}water & 31 & 45\\
            \rowfont{\color{cmapillary}}banner & 32 & 46\\
            \rowfont{\color{cmapillary}}bench & 33 & 47\\
            \rowfont{\color{cmapillary}}bike rack & 34 & 48\\
            \rowfont{\color{cmapillary}}billboard & 35 & 49\\
            \rowfont{\color{cmapillary}}catch basin & 36 & 50\\
            \rowfont{\color{cmapillary}}cctv camera & 37 & 51\\
            \rowfont{\color{cmapillary}}fire hydrant & 38 & 52\\
            \rowfont{\color{cmapillary}}junction box & 39 & 53\\
            \rowfont{\color{cmapillary}}mailbox & 40 & 54\\
            \rowfont{\color{cmapillary}}manhole & 41 & 55\\
            \rowfont{\color{cmapillary}}phone booth & 42 & 56\\
            \rowfont{\color{cmapillary}}pothole & 43 & 57\\
            street light & 44 & 17\\
            pole & 45 & 17\\
            traffic sign frame & 46 & 17\\
            utility pole & 47 & 17\\
            traffic light & 48 & 19\\
            traffic sign (back) & 49 & 20\\
            traffic sign (front) & 50 & 20\\
            \rowfont{\color{cmapillary}}trash can & 51 & 58\\
            bicycle & 52 & 33\\
            \rowfont{\color{cmapillary}}boat & 53 & 59\\
            bus & 54 & 28\\
            car & 55 & 26\\
            caravan & 56 & 29\\
            motorcycle & 57 & 32\\
            \rowfont{\color{cmapillary}}on rails & 58 & 60\\
            \rowfont{\color{cmapillary}}other vehicle & 59 & 61\\
            trailer & 60 & 30\\
            truck & 61 & 27\\
            \rowfont{\color{cmapillary}}wheeled slow & 62 & 62\\
            \rowfont{\color{cmapillary}}car mount & 63 & 63\\
            ego vehicle & 64 & 1\\
            unlabeled & -1 & 0\\
            \midrule
            \rowfont{\color{cnot_mapillary}}rectification border&-1&2\\
            \rowfont{\color{cnot_mapillary}}out of roi&-1&3\\
            \rowfont{\color{cnot_mapillary}}static&-1&4\\
            \rowfont{\color{cnot_mapillary}}dynamic&-1&5\\
            \rowfont{\color{cnot_mapillary}}ground&-1&6\\
            \rowfont{\color{cnot_mapillary}}polegroup&-1&18\\
            \rowfont{\color{cnot_mapillary}}train&-1&31\\
            \rowfont{\color{cnot_mapillary}}license plate&-1&34\\
            \bottomrule
            \multicolumn{3}{c}{\scriptsize \textbf{ (c) Mapillary semantic classes.}}
        \end{tabu}
        \vspace{0pt}
    \end{minipage}
    
    \label{tbl:csg0_mappings}
\end{table*}

%% file: main_arxiv_clvision.bbl
\begin{thebibliography}{10}\itemsep=-1pt

\bibitem{cong2020gan}
Yulai Cong, Miaoyun Zhao, Jianqiao Li, Sijia Wang, and Lawrence Carin.
\newblock Gan memory with no forgetting.
\newblock {\em NeurIPS}, 2020.

\bibitem{cordts2016cityscapes}
Marius Cordts, Mohamed Omran, Sebastian Ramos, Timo Rehfeld, Markus Enzweiler,
  Rodrigo Benenson, Uwe Franke, Stefan Roth, and Bernt Schiele.
\newblock The cityscapes dataset for semantic urban scene understanding.
\newblock In {\em CVPR}, 2016.

\bibitem{dumoulin2016learned}
Vincent Dumoulin, Jonathon Shlens, and Manjunath Kudlur.
\newblock A learned representation for artistic style.
\newblock {\em ICLR}, 2017.

\bibitem{heusel2017gans}
Martin Heusel, Hubert Ramsauer, Thomas Unterthiner, Bernhard Nessler, and Sepp
  Hochreiter.
\newblock Gans trained by a two time-scale update rule converge to a local nash
  equilibrium.
\newblock {\em NeuRIPS}, 2017.

\bibitem{huang2017arbitrary}
Xun Huang and Serge Belongie.
\newblock Arbitrary style transfer in real-time with adaptive instance
  normalization.
\newblock In {\em CVPR}, 2017.

\bibitem{pix2pix_isola2017image}
Phillip Isola, Jun-Yan Zhu, Tinghui Zhou, and Alexei~A Efros.
\newblock Image-to-image translation with conditional adversarial networks.
\newblock In {\em CVPR}, 2017.

\bibitem{kirkpatrick2017overcoming}
James Kirkpatrick, Razvan Pascanu, Neil Rabinowitz, Joel Veness, Guillaume
  Desjardins, Andrei~A Rusu, Kieran Milan, John Quan, Tiago Ramalho, Agnieszka
  Grabska-Barwinska, et~al.
\newblock Overcoming catastrophic forgetting in neural networks.
\newblock {\em PNAS}, 2017.

\bibitem{liu2019learning}
Xihui Liu, Guojun Yin, Jing Shao, Xiaogang Wang, et~al.
\newblock Learning to predict layout-to-image conditional convolutions for
  semantic image synthesis.
\newblock {\em NeurIPS}, 2019.

\bibitem{mo2020freeze}
Sangwoo Mo, Minsu Cho, and Jinwoo Shin.
\newblock Freeze the discriminator: a simple baseline for fine-tuning gans.
\newblock In {\em CVPR AI for Content Creation Workshop}, 2020.

\bibitem{neuhold2017mapillary}
Gerhard Neuhold, Tobias Ollmann, Samuel Rota~Bulo, and Peter Kontschieder.
\newblock The mapillary vistas dataset for semantic understanding of street
  scenes.
\newblock In {\em CVPR}, 2017.

\bibitem{noguchi2019image}
Atsuhiro Noguchi and Tatsuya Harada.
\newblock Image generation from small datasets via batch statistics adaptation.
\newblock In {\em ICCV}, 2019.

\bibitem{spade_park2019semantic}
Taesung Park, Ming-Yu Liu, Ting-Chun Wang, and Jun-Yan Zhu.
\newblock Semantic image synthesis with spatially-adaptive normalization.
\newblock In {\em CVPR}, 2019.

\bibitem{richter2016playing}
Stephan~R Richter, Vibhav Vineet, Stefan Roth, and Vladlen Koltun.
\newblock Playing for data: Ground truth from computer games.
\newblock In {\em ECCV}, 2016.

\bibitem{oasis_schonfeld2020you}
Edgar Sch{\"o}nfeld, Vadim Sushko, Dan Zhang, Juergen Gall, Bernt Schiele, and
  Anna Khoreva.
\newblock You only need adversarial supervision for semantic image synthesis.
\newblock In {\em ICLR}, 2020.

\bibitem{seff2017continual}
Ari Seff, Alex Beatson, Daniel Suo, and Han Liu.
\newblock Continual learning in generative adversarial nets.
\newblock {\em NeuRIPS}, 2017.

\bibitem{shahbazi2021efficient}
Mohamad Shahbazi, Zhiwu Huang, Danda~Pani Paudel, Ajad Chhatkuli, and Luc
  Van~Gool.
\newblock Efficient conditional gan transfer with knowledge propagation across
  classes.
\newblock In {\em CVPR}, 2021.

\bibitem{ulyanov2016instance}
Dmitry Ulyanov, Andrea Vedaldi, and Victor Lempitsky.
\newblock Instance normalization: The missing ingredient for fast stylization.
\newblock {\em arXiv preprint}, 2016.

\bibitem{varma2019idd}
Girish Varma, Anbumani Subramanian, Anoop Namboodiri, Manmohan Chandraker, and
  CV Jawahar.
\newblock Idd: A dataset for exploring problems of autonomous navigation in
  unconstrained environments.
\newblock In {\em WACV}, 2019.

\bibitem{pix2pixhd_wang2018high}
Ting-Chun Wang, Ming-Yu Liu, Jun-Yan Zhu, Andrew Tao, Jan Kautz, and Bryan
  Catanzaro.
\newblock High-resolution image synthesis and semantic manipulation with
  conditional gans.
\newblock In {\em CVPR}, 2018.

\bibitem{wang2020minegan}
Yaxing Wang, Abel Gonzalez-Garcia, David Berga, Luis Herranz, Fahad~Shahbaz
  Khan, and Joost van~de Weijer.
\newblock Minegan: effective knowledge transfer from gans to target domains
  with few images.
\newblock In {\em CVPR}, 2020.

\bibitem{wang2018transferring}
Yaxing Wang, Chenshen Wu, Luis Herranz, Joost van~de Weijer, Abel
  Gonzalez-Garcia, and Bogdan Raducanu.
\newblock Transferring gans: generating images from limited data.
\newblock In {\em ECCV}, 2018.

\bibitem{wu2018memory}
Chenshen Wu, Luis Herranz, Xialei Liu, Yaxing Wang, Joost Van~de Weijer, and
  Bogdan Raducanu.
\newblock Memory replay gans: learning to generate images from new categories
  without forgetting.
\newblock {\em NeuRIPS}, 2018.

\bibitem{zhai2020piggyback}
Mengyao Zhai, Lei Chen, Jiawei He, Megha Nawhal, Frederick Tung, and Greg Mori.
\newblock Piggyback gan: Efficient lifelong learning for image conditioned
  generation.
\newblock In {\em ECCV}, 2020.

\bibitem{zhai2021hyper}
Mengyao Zhai, Lei Chen, and Greg Mori.
\newblock Hyper-lifelonggan: Scalable lifelong learning for image conditioned
  generation.
\newblock In {\em CVPR}, 2021.

\bibitem{zhai2019lifelong}
Mengyao Zhai, Lei Chen, Frederick Tung, Jiawei He, Megha Nawhal, and Greg Mori.
\newblock Lifelong gan: Continual learning for conditional image generation.
\newblock In {\em ICCV}, 2019.

\bibitem{zhao2017pyramid}
Hengshuang Zhao, Jianping Shi, Xiaojuan Qi, Xiaogang Wang, and Jiaya Jia.
\newblock Pyramid scene parsing network.
\newblock In {\em CVPR}, 2017.

\end{thebibliography}
